\documentclass{article}

\usepackage{microtype}
\usepackage{graphicx}
\usepackage{booktabs} 

\usepackage{hyperref}

\usepackage[accepted]{icml2019}

\usepackage{url}
\usepackage{tikz}
\usepackage{float}
\usepackage{graphicx}
\usepackage{subcaption}
\usepackage{amsmath}
\usepackage{hyperref}
\usepackage{multirow}
\usepackage{adjustbox}
\usepackage{ragged2e} 
\usepackage{bm}
\usepackage{amsfonts}       

\usetikzlibrary{arrows}
\usetikzlibrary{positioning}
\usetikzlibrary{decorations.pathreplacing}
\makeatletter
    \let\pgf@decorate@@brace@brace@code@old\pgf@decorate@@brace@brace@code
    \def\pgf@decorate@@brace@brace@code{
        \ifdim\pgfdecoratedremainingdistance<4\pgfdecorationsegmentamplitude
            \pgftransformxscale{\pgfdecoratedremainingdistance/4\pgfdecorationsegmentamplitude}
            \pgfdecoratedremainingdistance=4\pgfdecorationsegmentamplitude
        \fi
        \pgf@decorate@@brace@brace@code@old
    }
\makeatother

\DeclareMathAlphabet\mathbfcal{OMS}{cmsy}{b}{n}

\icmltitlerunning{Multi-objective training of GANs with multiple discriminators}

\begin{document}

\twocolumn[
\icmltitle{Multi-objective training of Generative Adversarial Networks with multiple discriminators}



\icmlsetsymbol{equal}{*}

\begin{icmlauthorlist}
\icmlauthor{Isabela Albuquerque}{equal,inrs}
\icmlauthor{Jo\~ao Monteiro}{equal,inrs}
\icmlauthor{Thang Doan}{mcgill}
\icmlauthor{Breandan Considine}{mila}
\icmlauthor{Tiago Falk}{inrs}
\icmlauthor{Ioannis Mitliagkas}{mila}
\end{icmlauthorlist}

\icmlaffiliation{inrs}{INRS-EMT, Universit\'e du Qu\'ebec, Montreal, Canada}
\icmlaffiliation{mcgill}{Desautels Faculty of Management, McGill University, Montreal, Canada}
\icmlaffiliation{mila}{Quebec Artificial Intelligence Institute, Universit\'e de Montr\'eal, Montreal, Canada}

\icmlcorrespondingauthor{Isabela Albuquerque}{isabelamcalbuquerque@gmail.com}
\icmlcorrespondingauthor{Jo\~ao Monteiro}{joaomonteirof@gmail.com}

\icmlkeywords{Machine Learning, ICML}

\vskip 0.3in
]



\printAffiliationsAndNotice{\icmlEqualContribution} 

\begin{abstract}
Recent literature has demonstrated promising results for training Generative Adversarial Networks by employing a set of discriminators, in contrast to the traditional game involving one generator against a single adversary.
Such methods perform single-objective optimization on some simple consolidation of the losses, e.g. an arithmetic average.
In this work, we revisit the multiple-discriminator setting by framing the simultaneous minimization of losses provided by different models as a multi-objective optimization problem.
Specifically, we evaluate the performance of multiple gradient descent and the hypervolume maximization algorithm on a number of different datasets.
Moreover, we argue that the previously proposed methods and hypervolume maximization can all be seen as variations of multiple gradient descent in which the update direction can be computed efficiently.
Our results indicate that hypervolume maximization presents a better compromise between sample quality and computational cost than previous methods.
\end{abstract}

\section{Introduction}
Generative Adversarial Networks (GANs) \citep{goodfellow2014generative} offer a new approach to generative modeling, using game-theoretic training schemes to implicitly learn a given probability density. Prior to the emergence of GAN architectures, realistic generative modeling remained elusive. While offering unprecedented realism, GAN training still remains fraught with stability issues. Commonly reported shortcomings involve the lack of useful gradient signal provided by the discriminator, and mode collapse, i.e. lack of diversity in the generator's samples.

Considerable research effort has been devoted in recent literature to overcome training instability \footnote{\textit{Instability} in the sense commonly used in GANs literature, i.e. the discriminator is able to easily distinguish between real and fake samples during the training phase \citep{neyshabur2017stabilizing, arjovsky2017wasserstein, began}.} within the GAN framework. Some architectures such as BEGAN \citep{began} have applied auto-encoders as discriminators and proposed a new loss function to help stabilize training. Methods such as TTUR \citep{heusel2017gans}, in turn, have attempted to define separate schedules for updating the generator and discriminator. The PacGAN algorithm \citep{lin2017pacgan} proposes to modify the discriminator's architecture to accept \emph{m} concatenated samples as input. These samples are jointly classified as either real or generated, and the authors show that such an approach can help enforce sample diversity. Furthermore, \textit{spectral normalization} was introduced to the discriminator's parameters in SNGAN \citep{miyato2018spectral} aiming to ensure Lipschitz continuity, which is empirically shown to yield high quality samples across several sets of hyperparameters. Alternatively, recent works have proposed to tackle GANs instability issues with multiple discriminators. \citet{neyshabur2017stabilizing} propose a GAN variation in which one generator is trained against a set of discriminators, where each one sees a fixed random projection of the inputs. Prior work, including \citep{durugkar2016generative,acGAN} have also explored training with multiple discriminators.

In this paper, we build upon \citet{neyshabur2017stabilizing}'s introduced framework and propose reformulating the average loss minimization to further stabilize GAN training. 
Specifically, we propose treating the loss signal provided by each discriminator as an independent objective function. To achieve this, we simultaneously minimize the losses using multi-objective optimization techniques. Namely, we exploit previously introduced methods in literature such as the multiple gradient descent (MGD) algorithm \citep{desideri2012a}. However, due to MGD's prohibitively high cost in the case of large neural networks, we propose to use more efficient alternatives such as maximization of the hypervolume in the region defined between a fixed, shared upper bound on the losses, which we will refer to as the \textit{nadir point} $\boldsymbol{\eta}^*$, and each of the component losses.

In contrast to \citet{neyshabur2017stabilizing}'s approach, where the average loss is minimized when training the generator, hypervolume maximization (HV) optimizes a weighted loss, and the generator's training will adaptively assign greater importance to feedback from discriminators against which it performs poorly.

Experiments performed on MNIST show that HV presents a good compromise in the \emph{computational cost} vs. \emph{samples quality} trade-off, when compared to average loss minimization or GMAN's approach (low quality and cost), and MGD (high quality and cost). Also, the sensitivity to introduced hyperparameters is studied and results indicate that increasing the number of discriminators consequently increases the generator's robustness along with sample quality and diversity. Experiments on CIFAR-10 indicate the method described produces higher quality generator samples in terms of quantitative evaluation. Moreover, image quality and sample diversity are once more shown to consistently improve as we increase the number of discriminators.

In summary, our main contributions are the following:

\begin{enumerate}
\item We offer a new perspective on multiple-discriminator GAN training by framing it in the context of multi-objective optimization, and draw similarities between previous research in GANs variations and MGD, commonly employed as a general solver for multi-objective optimization.
\item We propose a new method for training multiple-discriminator GANs: Hypervolume maximization, which weighs the gradient contributions of each discriminator by its loss.
\end{enumerate}

The remainder of this document is organized as follows: Section \ref{sec: preliminaries} introduces definitions on multi-objective optimization and MGD. In Section \ref{sec: related_work} we describe prior relevant literature. Hypervolume maximization is detailed in Section \ref{sec: method}, with experiments and results presented in Section \ref{sec: experiments}. Conclusions and directions for future work are drawn in Section \ref{sec: conclusion}.

\section{Preliminaries}
\label{sec: preliminaries}

In this section we provide some definitions regarding multi-objective optimization from prior literature which will be useful in the following sections. Boldface notation is used to denote vector-valued variables.

\textbf{Multi-objective optimization.} A multi-objective optimization problem is defined as \citep{deb2001multi}:
\begin{equation}\label{MOP}
\begin{split}
\min \; & \textbf{F}(\textbf{x}) = [f_1(\textbf{x}), f_2(\textbf{x}), ... , f_K(\textbf{x})]^T, \\
&\textbf{x} \in \Omega,					
\end{split}
\end{equation}
where $K$ is the number of objectives, $\Omega$ is the variables space and $\textbf{x} = [x_1, x_2, ..., x_n]^T \in \Omega$ is a decision vector or possible solution to the problem. $\textbf{F}: \Omega \rightarrow \mathbb{R}^K$ is a set of $K$-objective functions that maps the $n$-dimensional variables space to the $K$-dimensional objective space.

\textbf{Pareto-dominance.} Let $\textbf{x}_1$ and $\textbf{x}_2$ be two decision vectors. $\textbf{x}_1$ is said to dominate $\textbf{x}_2$ (denoted by $\textbf{x}_1 \prec \textbf{x}_2$) if and only if  $f_i(\textbf{x}_1) \leq f_i(\textbf{x}_2)$ for all $i \in \{ 1,2,\ldots,K \}$ and $f_j(\textbf{x}_1) < f_j(\textbf{x}_2)$ for some $j \in \{ 1,2,\ldots,K \}$. If a decision vector $\textbf{x}$ is dominated by no other vector in $\Omega$, $\textbf{x}$ is called a non-dominated solution.

\textbf{Pareto-optimality.} A decision vector $\textbf{x}^* \in \Omega$ is said to be Pareto-optimal if and only if there is no $\textbf{x} \in \Omega$ such that $\textbf{x} \prec \textbf{x}^*$, i.e. $\textbf{x}^*$ is a non-dominated solution. The Pareto-optimal Set (PS) is defined as the set of all Pareto-optimal solutions $\textbf{x} \in \Omega$, i.e., $PS = \{ \textbf{x} \in \Omega | \text{\textbf{x} is Pareto optimal} \}$. The set of all objective vectors $\textbf{F}(\textbf{x})$ such that $\textbf{x}$ is Pareto-optimal is called Pareto front (PF), that is $PF = \{\textbf{F}(\textbf{x}) \in \mathbb{R}^K | \textbf{x} \in PS \}$.

\textbf{Pareto-stationarity.}  Pareto-stationarity is a necessary condition for Pareto-optimality. For $f_k$ differentiable everywhere for all $k$, $\textbf{F}$ is Pareto-stationary at $\textbf{x}$ if there exists a set of scalars $\alpha_k, k \in \{1, \ldots, K\}$, such that:
\begin{equation}
\label{eq: pareto_stationary}
\sum_{k=1}^K \alpha_k \nabla f_k = \textbf{0}, \quad \sum_{k=1}^K \alpha_k = 1, \quad \alpha_k \geq 0  \quad \forall k.
\end{equation}
\textbf{Multiple Gradient Descent.} Multiple gradient descent \citep{desideri2012a, schaffler2002stochastic,peitz2018gradient} was proposed for the unconstrained case of multi-objective optimization of $\textbf{F}(\textbf{x})$ assuming a convex, continuously differentiable and smooth $f_k(\textbf{x})$ for all $k$. MGD finds a common descent direction for all $f_k$ by defining the convex hull of all $\nabla f_k(\textbf{x})$ and finding the minimum norm element within it. Consider  $\textbf{w}^*$ given by:
\begin{equation}\label{eq: descent_dir}
\begin{split}
& \textbf{w}^*=\text{argmin} ||\textbf{w}||, \quad \textbf{w} = \sum_{k=1}^K \alpha_k \nabla f_k(\textbf{x}), \\
& \quad \text{s.t. } \quad \sum_{k=1}^K \alpha_k = 1, \quad \alpha_k \geq 0  \quad \forall k.   
\end{split}
\end{equation}
$\textbf{w}^*$ will be either $\textbf{0}$ in which case $\textbf{x}$ is a Pareto-stationary point, or $\textbf{w}^* \neq \textbf{0}$ and then $\textbf{w}^*$ is a descent direction for all $f_i(\textbf{x})$. Similar to gradient descent, MGD consists in finding the \textit{common} steepest descent direction $\textbf{w}^*_t$ at each iteration $t$, and then updating parameters with a learning rate $\lambda$ according to $\textbf{x}_{t+1}=\textbf{x}_{t} - \lambda \frac{\textbf{w}^*_t}{||\textbf{w}^*_t||}$.

\section{Related work}
\label{sec: related_work}

\subsection{Training GANs with multiple discriminators}
While we would prefer to always have strong gradients from the discriminator during training, the vanilla GAN makes this difficult to ensure, as the discriminator quickly learns to distinguish real and generated samples \citep{goodfellow2016nips}, thus providing no meaningful error signal to improve the generator thereafter. \citet{durugkar2016generative} proposed the Generative Multi-Adversarial Networks (GMAN) which consists of training the generator against a \textit{softmax} weighted arithmetic average of $K$ different discriminators: 
\begin{equation}\label{eq:l_gman}
\mathcal{L}_G = \sum_{k=1}^K \alpha_k \mathcal{L}_{D_k},
\end{equation}
where $\alpha_k = \frac{e^{\beta\mathcal{L}_{D_k}}}{\sum_{j=1}^K e^{\beta\mathcal{L}_{D_j}}}$, $\beta \geq 0$, and $\mathcal{L}_{D_k}$ is the loss of discriminator $k$ and is defined as 
\begin{equation}\label{eq:k-disc}
\mathcal{L}_{D_k} = -\mathbb{E}_{\textbf{x} \sim p_{\text{data}}} \log D_k(\textbf{x}) -\mathbb{E}_{\textbf{z} \sim p_{z}} \log (1 - D_k(G(\textbf{z}))),
\end{equation}
where $D_k(\textbf{x})$ and $G(\textbf{z})$ are the outputs of the $k$-th discriminator and the generator, respectively.
The goal of using the proposed averaging scheme is to favor worse discriminators, thus providing more useful gradients to the generator during training. Experiments were performed with $\beta = 0$ (equal weights), $\beta \rightarrow \infty$ (only worst discriminator is taken into account), $\beta = 1$, and $\beta$ learned by the generator. Models with $K=\{2, 5\}$ were tested and evaluated using a proposed metric and the Inception score \citep{salimans2016improved}. Results showed that the simple average of discriminator's losses provided the best values for both metrics in most of the considered cases.


\citet{neyshabur2017stabilizing} proposed training a GAN with $K$ discriminators using the same architecture. Each discriminator $D_k$ sees a different randomly projected lower-dimensional version of the input image. Random projections are defined by a randomly initialized matrix $W_k$, which remains fixed during training. Theoretical results provided show the distribution induced by the generator $G$ will converge to the real data distribution $p_{\text{data}}$, as long as there is a sufficient number of discriminators. Moreover, discriminative tasks in the projected space are harder, i.e. real and fake examples are more alike, thus avoiding early convergence of discriminators, which leads to common stability issues in GAN training such as mode-collapse \citep{goodfellow2016nips}. Essentially, the authors trade one hard problem for $K$ easier subproblems. The losses of each discriminator $\mathcal{L}_{D_k}$ are the same as shown in Eq. \ref{eq:k-disc}. However, the generator loss $\mathcal{L}_G$ is defined as the sum of the losses provided by each discriminator, as shown in Eq. \ref{eq:gen}. This choice of $\mathcal{L}_G$ does not exploit available information such as the performance of the generator with respect to each discriminator.
\begin{equation}\label{eq:gen}
\mathcal{L}_G = -\sum_{k=1}^K \mathbb{E}_{\textbf{z} \sim p_{z}} \log D_k(G(\textbf{z})). 
\end{equation}
\subsection{Hypervolume maximization}

Let $S$ be the solutions for a multi-objective optimization problem. The hypervolume $\mathcal{H}$ of $S$ is defined as \citep{fleischer2003measure}: $\mathcal{H}(S) = \mu( \cup_{\textbf{x}\in S} [\textbf{F}(\textbf{x}), \boldsymbol{\eta}^*])$, where $\mu$ is the Lebesgue measure and $\boldsymbol{\eta}^*$ is a point dominated by all $\textbf{x}\in S$ (i.e. $f_i(\textbf{x})$ is upper-bounded by $\eta$), referred to as the \textit{nadir point}. $\mathcal{H}(S)$ can be understood as the size of the space covered by $\{\textbf{F}(\textbf{x}) | \textbf{x} \in S\}$ \citep{bader2011hype}. 

The hypervolume was originally introduced as a quantitative metric for coverage and convergence of Pareto-optimal fronts obtained through population-based algorithms \citep{beume2007sms}. Methods based on direct maximization of $\mathcal{H}$ exhibit favorable convergence even in challenging scenarios, such as simultaneous minimization of 50 objectives \citep{bader2011hype}. In the context of Machine Learning, single-solution hypervolume maximization has been applied to neural networks as a surrogate loss for mean squared error \citep{MirandaZ16}, i.e. the loss provided by each example in a training batch is treated as a single cost and the multi-objective approach aims to minimize costs over all examples. Authors show that such method provides an inexpensive boosting-like training.

\section{Multi-objective training of GANs with multiple discriminators}
\label{sec: method}

We introduce a variation of the GAN game in which the generator solves the following multi-objective problem:
\begin{equation}\label{MOGAN}
\min \mathbfcal{L}_G(\textbf{x}) = [ l_1(\textbf{z}), l_2(\textbf{z}), ... , l_K(\textbf{z}) ]^T,
\end{equation}
where each $l_k = -\mathbb{E}_{z \sim p_{z} } \log D_k(G(z))$, $k \in \{1,...,K\}$, is the loss provided by the $k$-th discriminator. Training proceeds in the usual fashion \citep{goodfellow2014generative}, i.e. with alternate updates between the discriminators and the generator. Updates of each discriminator are performed to minimize the loss described in Eq. \ref{eq:k-disc}.

A natural choice for our generator's updates is the MGD algorithm, described in Section \ref{sec: preliminaries}. However, computing the direction of steepest descent $\textbf{w}^*$ before every parameter update step, as required in MGD, can be prohibitively expensive for large neural networks. Therefore, we propose an alternative scheme for multi-objective optimization and argue that both our proposal and previously published methods can all be viewed as performing a computationally more efficient version of the MGD update rule, without the burden of needing to solve a quadratric program, i.e. computing $\textbf{w}^*$, every iteration.

\subsection{Hypervolume maximization for training GANs}
\begin{figure}[h]
\centering
  \begin{tikzpicture}[scale=2]
    \draw[->] (-0.1,0) -- (1.4, 0) node[right] {$\mathcal{L}_{D_1}$};
    \draw[->] (0,-0.1) -- (0, 1.2) node[above] {$\mathcal{L}_{D_2}$};
    \draw[line width=1pt] (0.6,-0.02) -- (0.6, 0.02) node[below=0.2cm] {$l_1$};
    \draw[dashed] (0.6,0.0) -- (0.6, 1.0);
    \draw[line width=1pt] (-0.02, 0.6) -- (0.02, 0.6) node[left=0.2cm] {$l_2$};
    \draw[dashed] (0.0, 0.6) -- (1.2, 0.6);
    \draw [fill] (1.2, 1.0) circle [radius=0.017] node[right=0.1cm] {$\boldsymbol{\eta}^*$};
    \draw [fill] (0.6, 0.6) circle [radius=0.017] node[left=0.22cm, below] {$\boldsymbol{l}$};
    \draw[dashed] (1.2, 0.0) -- (1.2, 1.0);
    \draw[line width=1pt] (1.2,-0.02) -- (1.2, 0.02) node[below=0.2cm] {$\eta$};
    \draw[dashed] (0.0, 1.0) -- (1.2, 1.0);
    \draw[line width=1pt] (-0.02, 1.0) -- (0.02, 1.0) node[left=0.2cm] {$\eta$};
    \fill[fill=pink, draw=none] (0.612,0.612) -- (0.612, 0.992) -- (1.192,0.992) -- (1.192,0.612) node[midway,left=0.2cm] {$e^{\mathcal{V}}$};
  \end{tikzpicture}
\captionof{figure}{2D example of the objective space where the generator loss is being optimized.}
\label{fig:hypervolume}
\end{figure}
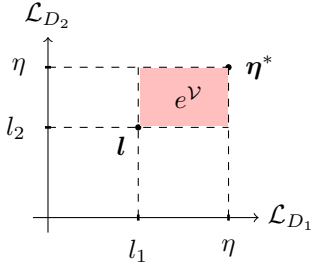

\citet{fleischer2003measure} has shown that maximizing $\mathcal{H}$ yields Pareto-optimal solutions. Since MGD converges to a set of Pareto-stationary points, i.e. a superset of the Pareto-optimal solutions, hypervolume maximization yields a subset of the solutions obtained using MGD. We exploit this property and define the generator loss as the negative log-hypervolume, as defined in Eq. \ref{eq:hypervolume}:
\begin{equation}
\label{eq:hypervolume}
\mathcal{L}_G = -\mathcal{V} = -\sum_{k=1}^{K} \log (\eta - l_k),
\end{equation}
where the nadir point coordinate $\eta$ is an upper bound for all $l_k$. In Fig. \ref{fig:hypervolume} we provide an illustrative example for the case where $K=2$. The highlighted region corresponds to $e^{\mathcal{V}}$. Since the nadir point $\boldsymbol{\eta}^*$ is fixed, $\mathcal{V}$ will be maximized, and consequently $\mathcal{L}_G$ minimized, if and only if each $l_k$ is minimized.
Moreover, by adapting the results shown in \citep{MirandaZ16}, the gradient of $\mathcal{L}_G$ with respect to any generator's parameter $\theta$ is given by:
\begin{equation}
\label{eq:grad_hv}
\frac{\partial \mathcal{L}_G}{\partial \theta} = \sum_{k=1}^K \frac{1}{\eta-l_k}\frac{\partial l_k}{\partial \theta}.
\end{equation}
In other words, the gradient can be obtained by computing a weighted sum of the gradients of the losses provided by each discriminator, whose weights are defined as the inverse distance to the nadir point components. This formulation will naturally assign more importance to higher losses in the final gradient, which is another useful property of hypervolume maximization.

\textbf{Nadir point selection.} It is evident from Eq. \ref{eq:grad_hv} that the selection of $\eta$ directly affects the importance assignment of gradients provided by different discriminators. Particularly, as the quantity \mbox{$\min_k \{ \eta - l_k \}$} grows, the multi-objective GAN game approaches the one defined by the simple average of $l_k$. Previous literature has discussed in depth the effects of the selection of $\eta$ in the case of population-based methods \citep{auger2009theory,auger2012hypervolume}. However, those results are not readily applicable for the single-solution case. As will be shown in Section \ref{sec: experiments}, our experiments indicate that the choice of $\eta$ plays an important role in the final quality of samples. Nevertheless, this effect becomes less relevant as the number of discriminators increases.

\textbf{Nadir point adaptation.} Similarly to \citep{MirandaZ16}, we propose an adaptive scheme for $\eta$ such that at iteration \mbox{$t$: $\eta^t = \delta \max_k \{ l_{k}^{t} \}$}, where $\delta>1$ is a user-defined parameter which will be referred to as \textit{slack}. This enforces $\min_k \{ \eta^t - l_k^t \}$ to be higher when $\max_k \{ l_{k}^{t} \}$ is high and low otherwise, which induces a similar behavior as an average loss when training begins and automatically places more importance on the discriminators in which performance is worse as training progresses.

\begin{figure}[h]
    \centering
    \begin{tikzpicture}[scale=2]
    \draw[->] (-0.1,0) -- (1.9, 0) node[right] {$\mathcal{L}_{D_1}$};
    \draw[->] (0,-0.1) -- (0, 1.8) node[above] {$\mathcal{L}_{D_2}$};
    \draw[line width=1pt] (0.4,-0.02) -- (0.4, 0.02) node[below=0.2cm] {$l_1^T$};
    \draw[dashed] (0.4, 0.0) -- (0.4, 0.8);
    \draw[line width=1pt] (-0.02, 0.3) -- (0.02, 0.3) node[left=0.2cm] {$l_2^T$};
    \draw[dashed] (0.0, 0.3) -- (0.8, 0.3);
    \draw [fill] (0.8, 0.8) circle [radius=0.017] node[right=0.1cm] {$\boldsymbol{\eta}^*$};
    \draw [fill] (0.4, 0.3) circle [radius=0.017] node[left=0.25cm, below = 0.08cm] {};
    \draw[dashed] (0.8, 0.0) -- (0.8, 0.8);
    \draw[line width=1pt] (0.8,-0.02) -- (0.8, 0.02) node[below=0.2cm] {$\eta^T$};
    \draw[dashed] (0.0, 0.8) -- (0.8, 0.8);
    \draw[line width=1pt] (-0.02, 0.8) -- (0.02, 0.8) node[left=0.2cm] {$\eta^T$};
    \fill[fill=pink, draw=none] (0.412,0.312) -- (0.412, 0.788) -- (0.788, 0.788) -- (0.788, 0.312) node[left=0.3cm, above=0.3cm] {};
    \draw [decorate, decoration={brace, amplitude=8pt, mirror}] (0.4, -0.4) -- (0.8, -0.4) node [black, midway, yshift=-0.55cm] {\footnotesize $\eta^T - \max\{l_1^T, l_2^T\}$};
    \draw[dashed, red] (1.7, 0.0) -- (1.7, 1.7) {};
    \draw[line width=1pt, red] (1.7,-0.02) -- (1.7, 0.02) node[below=0.2cm] {$\eta^0$};
    \draw[dashed, red] (0.0, 1.7) -- (1.7, 1.7);
    \draw[line width=1pt, red] (-0.02, 1.7) -- (0.02, 1.7) node[left=0.2cm] {$\eta^0$};
    \draw [fill, red] (1.7, 1.7) circle [radius=0.017] node[right=0.1cm] {$\boldsymbol{\eta}^*$};
    \draw [red, decorate, decoration={brace, amplitude=6pt}] (0.4, -0.01) -- (1.7, -0.01) node [red, midway, yshift=0.3cm, xshift=0.2cm] {\tiny $\eta^0 - \max\{l_1^T, l_2^T\}$};
  \end{tikzpicture}
  \caption{Losses and nadir point at $t=T$, and nadir point at $t=0$ (in red).}
  \label{fig:hvnadir_T}
\end{figure}
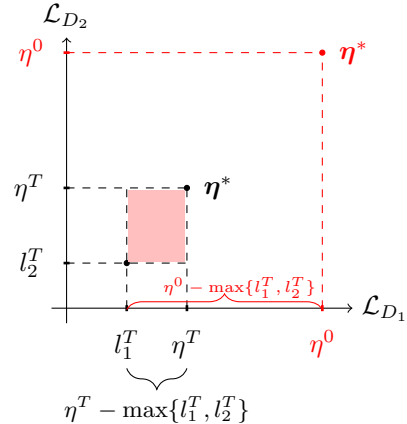

We further illustrate the proposed adaptation scheme in Fig. \ref{fig:hvnadir_T}. Consider a two-objective problem with $l_1^t>0$ and $l_2^t>0$ corresponding to $\mathcal{L}_{D_1}$ and $\mathcal{L}_{D_2}$ at iteration $t$, respectively. If no adaptation is performed and $\eta$ is left unchanged throughout training, as represented by the red dashed lines in Fig. \ref{fig:hvnadir_T}, $\eta - l_1^t \approx \eta - l_2^t$ for a large enough $t$. This will assign similar weights to gradients provided by the different losses, which defeats the purpose of employing hypervolume maximization rather than average loss minimization. Assuming that losses decrease with time, after $T$ updates, $\eta^T = \delta \max \{ l_1^T, l_2^T \} < \eta$ , since losses are now closer to $0$. The employed adaptation scheme thus keeps the gradient weighting relevant even when losses become low. This effect will become more aggressive as training progresses, assigning more gradient importance to higher losses, as $\eta^T - \max\{l_1^T, l_2^T\} < \eta^0 - \max\{l_1^0, l_2^0\}$.

\textbf{Comparison to average loss minimization.} The upper bound proven by \citet{neyshabur2017stabilizing} assumes that the marginals of the real and generated distributions are identical along all random projections. However, average loss minimization does not ensure equally good approximation between the marginals along all directions. 
In the case of competing discriminators, i.e. if decreasing the loss on a given projection increases the loss with respect to another one, the distribution of losses can be uneven.
With HV on the other hand, especially when $\eta$ is reduced throughout training, the overall loss will be kept high as long as there are discriminators with high loss.
This objective tends to prefer central regions, in which all discriminators present a roughly equally low loss.

\subsection{Relationship between multiple discriminator GANs and MGD}
All methods described previously for the solution of GANs with multiple discriminators, i.e. average loss minimization \citep{neyshabur2017stabilizing}, GMAN's weighted average \citep{durugkar2016generative} and hypervolume maximization can be defined as MGD-like two-step algorithms consisting of: \textit{Step 1} - consolidate all gradients into a single update direction (compute the set $\alpha_{1,...,K}$); \textit{Step 2} - update parameters in the direction returned in Step 1. The definition of \textit{Step 1} for the different methods studied here can be summarized as follows:
\begin{enumerate}
\item MGD: $\alpha_{1:K} = \text{argmin}_\alpha ||\textbf{w}||, \quad  $   s.t. $\quad \sum_{k=1}^K \alpha_k = 1$, $\quad \alpha_k \geq 0$   $\forall k \in \{1,...,K\}$
\item Average loss minimization \citep{neyshabur2017stabilizing}: $\alpha_k = \frac{1}{K}$
\item GMAN \citep{durugkar2016generative}: $\alpha_k = \text{softmax}(l_{1:K})_k$
\item Hypervolume maximization: $\alpha_k = \frac{1}{T(\eta-l_k)},$ 
\\ $T = \sum_{k=1}^K \frac{1}{\eta-l_k}$ 
\end{enumerate}
\section{Experiments}
\label{sec: experiments}

We performed four sets of experiments aiming to understand the following phenomena: (i) How alternative methods for training GANs with multiple discriminators perform in comparison to MGD; (ii) How alternative methods perform in comparison to each other in terms of sample quality and coverage; (iii) How the varying number of discriminators impacts performance given the studied methods; and (iv) Whether the multiple-discriminator setting is practical given the added cost involved in training a set of discriminators.

Firstly, we exploited the relatively low dimensionality of MNIST and used it as testbed for comparing MGD with the other approaches, i.e. average loss minimization (AVG), GMAN's weighted average loss, and HV, proposed in this work. Moreover, multiple initializations and \textit{slack} combinations were evaluated in order to investigate how varying the number of discriminators affects robustness to those factors.

Then, experiments were performed with an upscaled version of CIFAR-10 at the resolution of 64x64 pixels while increasing the number of discriminators. Upscaling was performed with the aim of running experiments utilizing the same architecture described in \cite{neyshabur2017stabilizing}. We evaluated HV's performance compared to baseline methods in terms of its resulting sample quality. Additional experiments were carried out with CIFAR-10 at its original resolution in order to provide a clear comparison with well known single-discriminator settings. We further analyzed HV's impact on the diversity of generated samples using the stacked MNIST dataset \citep{srivastava2017veegan}. Finally, the computational cost and performance are compared for the single- vs. multiple-discriminator cases. Samples of generators trained on stacked MNIST in the Appendix along with samples from CelebA at a $128 \times 128$ resolution as well as the Cats dataset at a $256 \times 256$ resolution.

In all experiments performed, the same architecture, set of hyperparameters and initialization were used for both AVG, GMAN and our proposed method, the only variation being the generator loss. Unless stated otherwise, Adam \citep{kingma2014adam} was used to train all the models with learning rate, $\beta_1$ and $\beta_2$ set to $0.0002$, $0.5$ and $0.999$, respectively. Mini-batch size was set to $64$. The Fr{\'e}chet Inception Distance (FID) \citep{heusel2017gans} was used for comparison. Details on FID computation can be found in Appendix A.

\subsection{MGD compared with alternative methods}

We employed MGD in our experiments with MNIST and, in order to do so, a quadratic program has to be solved prior to every parameters update. For this, we used Scipy's implementation of the Serial Least Square Quadratic Program solver\footnote{\url{https://docs.scipy.org/doc/scipy/reference/tutorial/optimize.html}}. Three and four fully connected layers with \textit{LeakyReLU} activations were used for the generator and discriminator, respectively. Dropout was also employed in the discriminator and the random projection layer was implemented as a randomly initialized $\text{norm-}1$ fully connected layer, reducing the vectorized dimensionality of MNIST from $784$ to $512$. The output layer of a pretrained \textit{LeNet} \citep{lecun1998gradient} was used for FID computation.

\begin{figure}[h]
\centering
\includegraphics[width=\columnwidth]{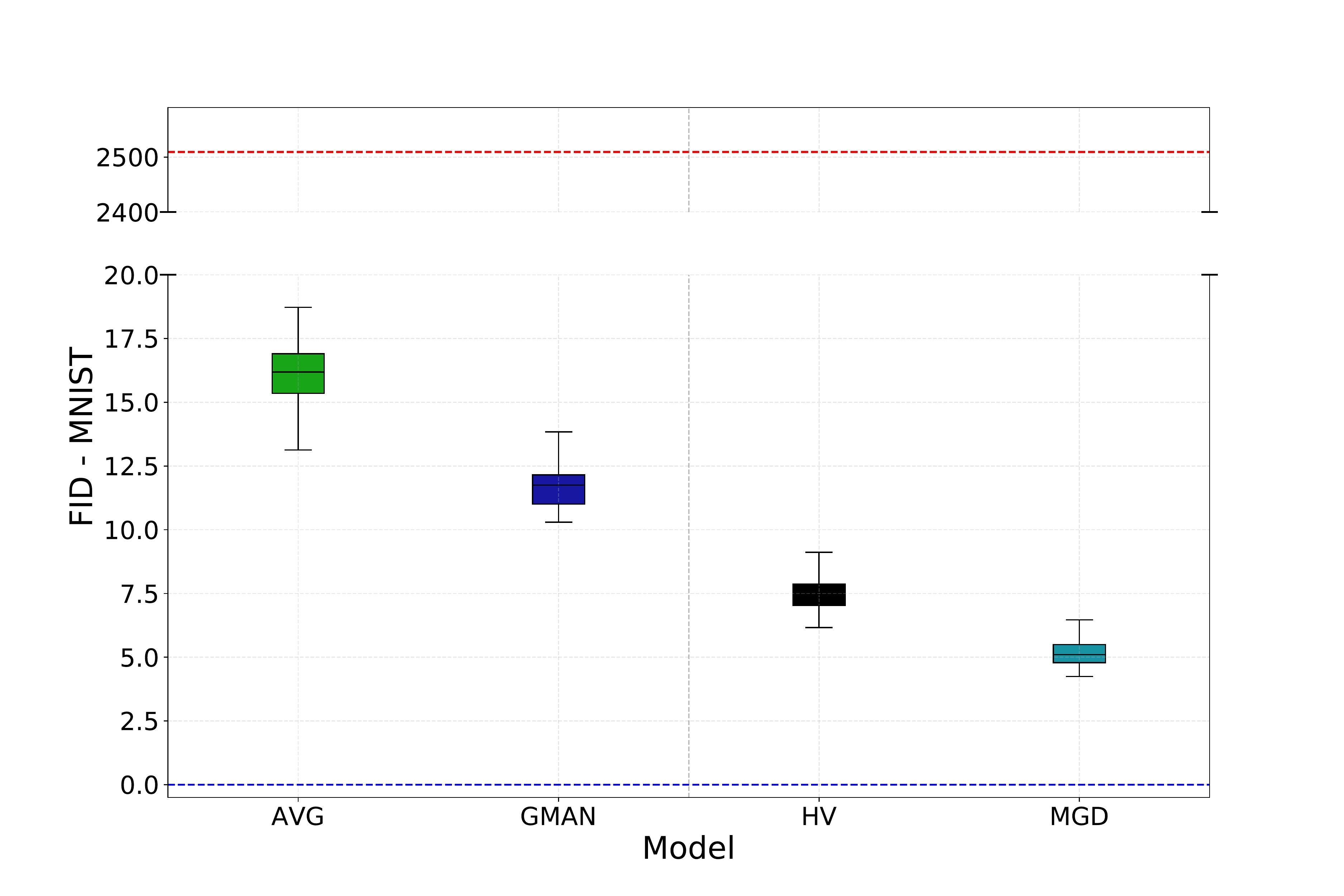}
\caption{Box-plots corresponding to 30 independent FID computations with $10000$ images. MGD performs consistently better than other methods, followed by hypervolume maximization. Models that achieved minimum FID at training time were used. Red and blue dashed lines represent FID values for a random generator and real data, respectively.}
\label{fig:FID_mnist}
\end{figure}

\begin{figure}[h]
\centering
\includegraphics[width=\columnwidth]{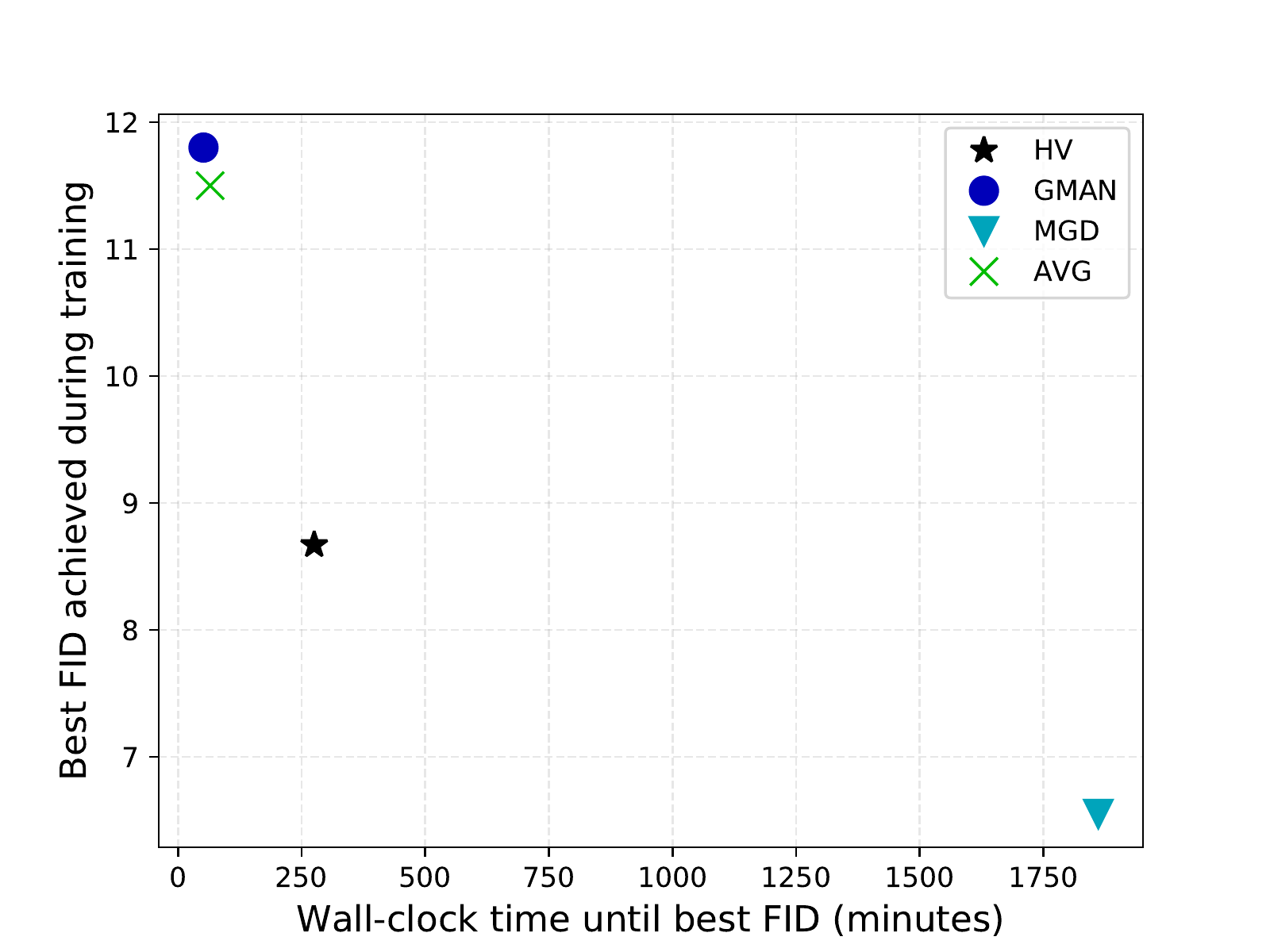}
\caption{Time vs. best FID achieved during training for each approach. FID values are computed over $1000$ generated images after every epoch. MGD performs relevantly better than others in terms of FID, followed by HV. However, MGD is approximately 7 times slower than HV. HV is well-placed in the time-quality trade-off.}
\label{fig:mnist}
\end{figure}

Experiments over $100$ epochs with $8$ discriminators are reported in Fig. \ref{fig:FID_mnist} and Fig. \ref{fig:mnist}. In Fig. \ref{fig:FID_mnist}, box-plots refer to $30$ independent computations of FID over $10000$ images sampled from the generator which achieved the minimum FID at train time. FID results are measured at training time with over $1000$ images and the best values are reported in Fig. \ref{fig:mnist} along with the necessary time to achieve it.

MGD outperforms all tested methods. However, its cost per iteration does not allow its use in more relevant datasets outside MNIST. Hypervolume maximization, on the other hand, performs closer to MGD than the considered baselines, while introducing no relevant extra cost.

\begin{figure}[h!]
\centering
\includegraphics[width=\columnwidth]{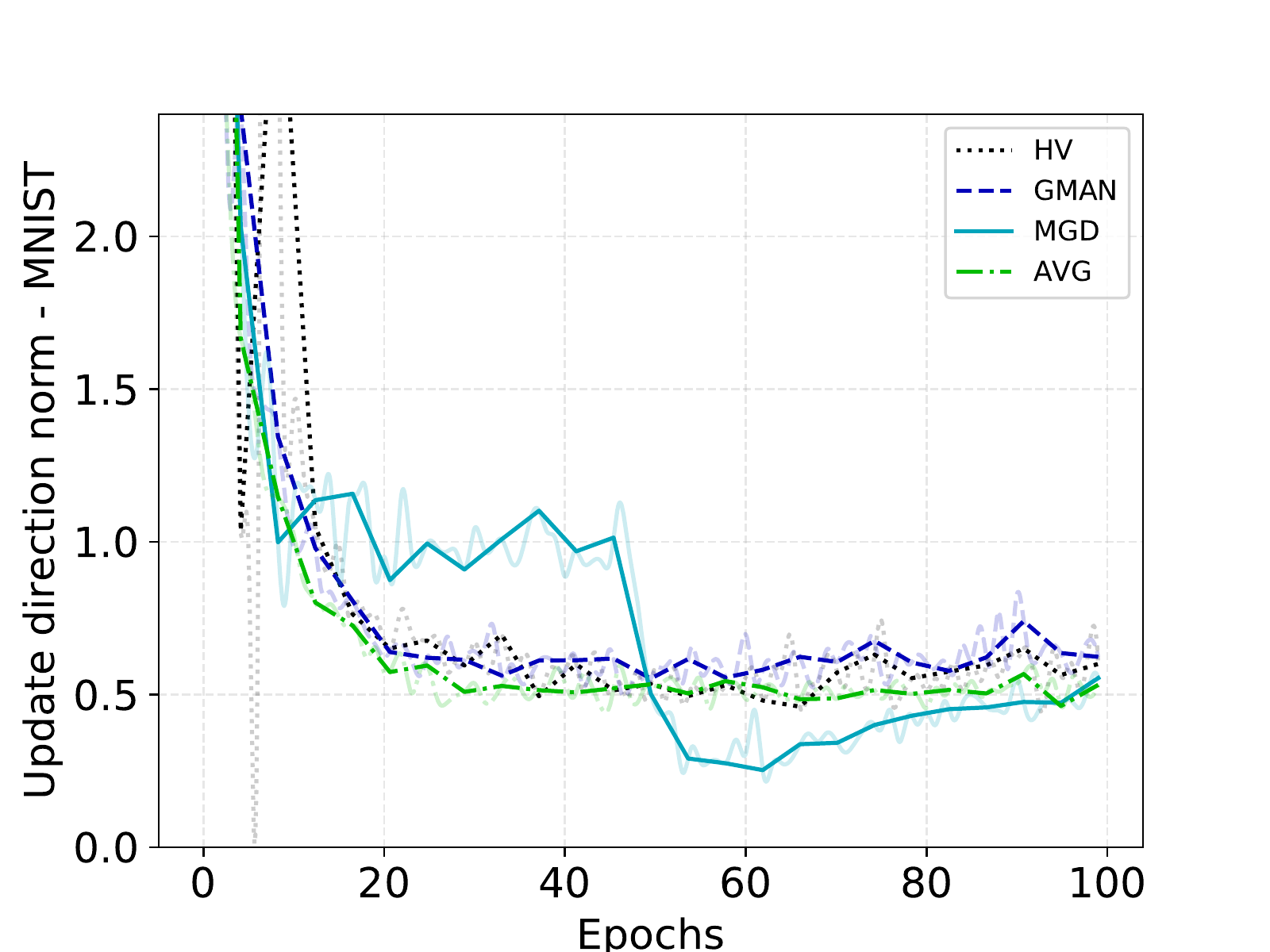}
\caption{Norm of the update direction over time for each method. While Pareto-stationarity is approximately achieved by all methods, performance varies relevantly in terms of FID.}
\label{fig:steep_mnist}
\end{figure}

\begin{figure}[h]
    \centering
    \includegraphics[width=.85\textwidth]{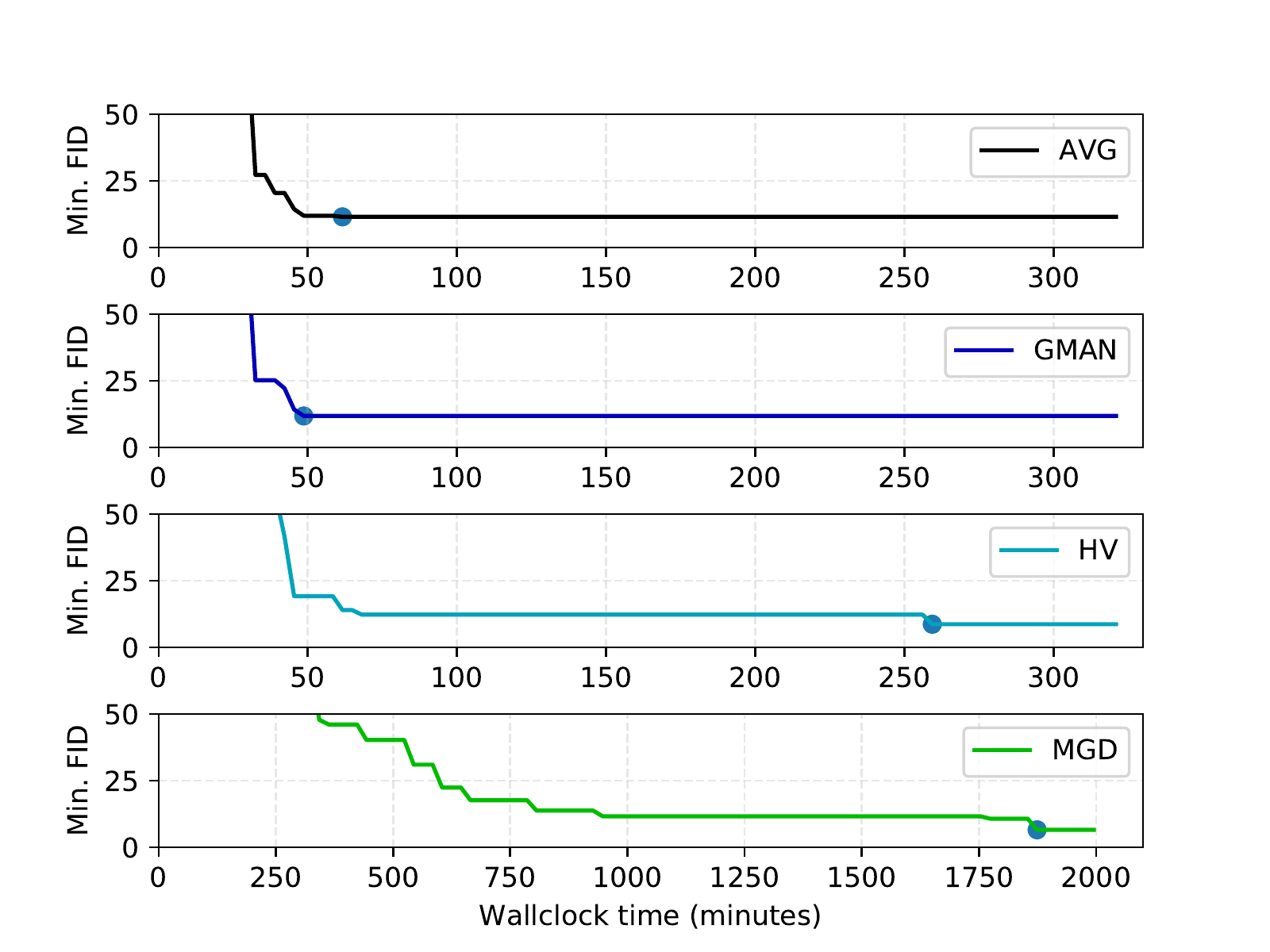}
    \caption{Minimum FID during training. X-axis is in minutes. The blue dot is intended to highlight the moment during training when the minimum FID was reached.}
    \label{fig:fid_time_mnist}
\end{figure}

In Fig. \ref{fig:steep_mnist}, we analyze convergence in the Pareto-stationarity sense, by plotting the norm of the update direction for each method, given by $||\sum_{k=1}^K \alpha_k \nabla l_k||$. All methods converged to similar norms, leading to the conclusion that different Pareto-stationary solutions will perform differently in terms of quality of samples. Best FID as a function of wall-clock time is shown in Fig. \ref{fig:fid_time_mnist} at the Appendix.

\begin{figure}[htbp]
\centering
\includegraphics[width=\columnwidth]{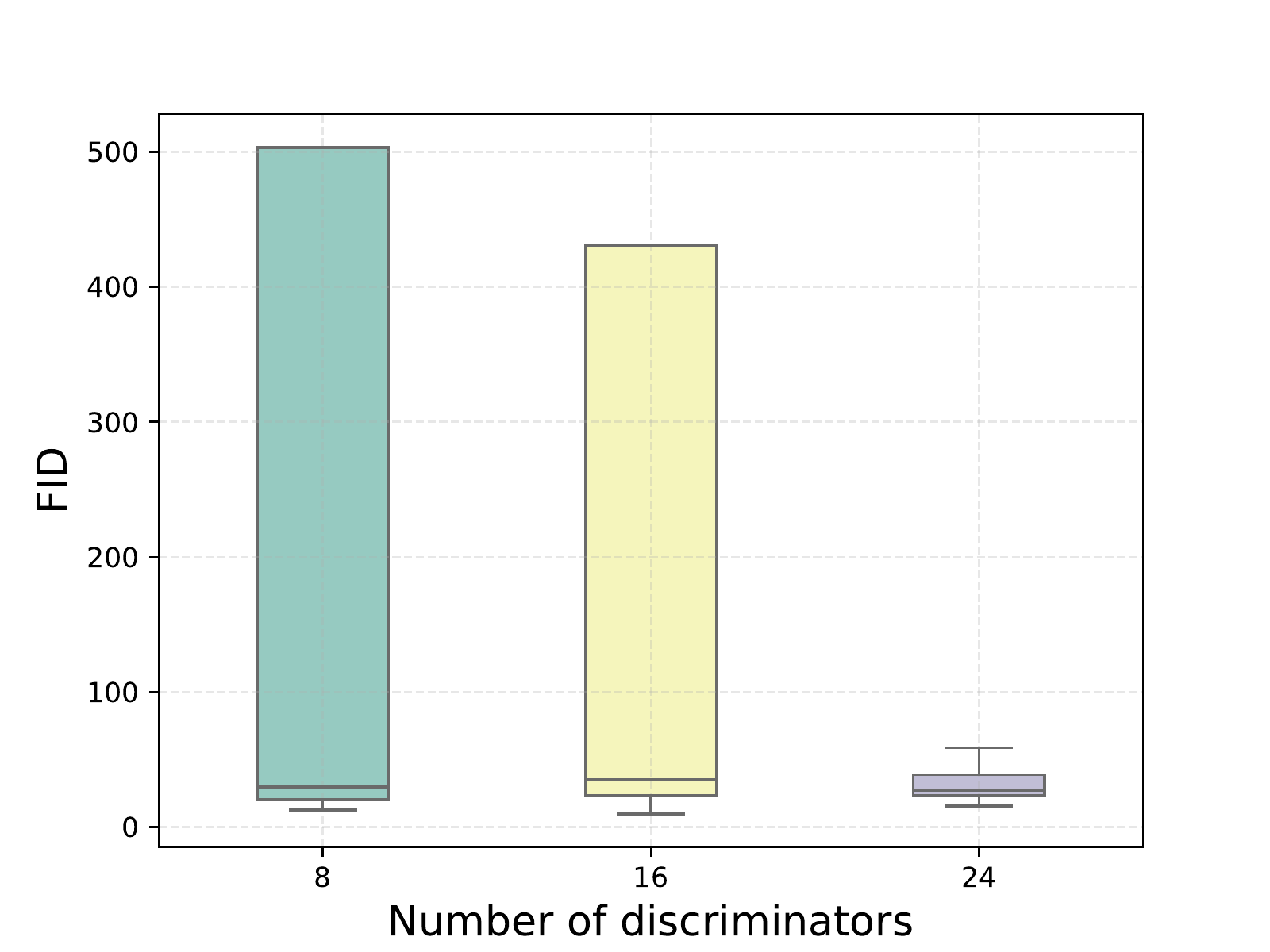}
\captionof{figure}{Independent FID evaluations for models obtained with different initializations and slack parameter $\delta$. Sensitivity reduces as the number of discriminators increases.}
\label{fig:fid_hyper_manydeltas}
\end{figure}

\textbf{HV sensitivity to initialization and choice of $\delta$}. Analysis of the performance sensitivity with the choice of the slack parameter $\delta$ and initialization was performed under the following setting: models were trained for $50$ epochs on MNIST with hypervolume maximization using 8, 16, 24 discriminators. Three independent runs (different initializations) were executed  with each $\delta=\{1.05, 1.5, 1.75, 2\}$ and number of discriminators, totaling 36 final models. Fig. \ref{fig:fid_hyper_manydeltas} reports the box-plots obtained for $5$ FID independent computations using $10000$ images, for each of the $36$ models obtained under the setting described. Results clearly indicate that increasing the number of discriminators yields much smaller variation in the FID obtained by the final model.

\subsection{HV as an alternative for MGD}
\subsubsection{Upscaled CIFAR-10}
\label{sec:up_cifar}
We evaluate the performance of HV compared to baseline methods using the upscaled CIFAR-10 dataset. FID was computed with a pretrained ResNet \citep{he2016deep}. ResNet was trained on the 10-class classification task of CIFAR-10 up to approximately $95 \%$ test accuracy. DCGAN \citep{radford2015unsupervised} and WGAN-GP \citep{gulrajani2017improved} were included in the experiments for FID reference. Same architectures as in \citep{neyshabur2017stabilizing} were employed for all multi-discriminators settings. An increasing number of discriminators was used. Inception score \citep{salimans2016improved} as well as FID computed with other models are included in the Appendix-Table \ref{tab:scores_cifar}.

\begin{table*}
\centering
\begin{tabular}{lccccc}
\hline
                                         & FID-ResNet & FID (5k) & IS (5k) & FID (10k) & IS (10k) \\ \hline
SNGAN \citep{miyato2018spectral}                      & -                & 25.5              & $7.58 \pm 0.12$      & -                  & -                     \\
WGAN-GP \citep{miyato2018spectral}                    & -                & 40.2              & $6.68 \pm 0.06$      & -                  & -                     \\
DCGAN \citep{miyato2018spectral}                      & -                & -                 & $6.64 \pm 0.14$      & -                  & -                     \\ \hline
SNGAN (our implementation)               & 1.55             & 27.93             & $7.11 \pm 0.30$      & 25.29      &   $7.26 \pm 0.12$                  \\
DCGAN + 24 Ds and HV   & 1.21             & 27.74           & $7.32 \pm 0.26$    & 24.90              &  $7.45\pm0.17$                   \\ \hline
\end{tabular}
\caption{Evaluation of the effect of adding discriminators on a DCGAN-like model trained on CIFAR-10. Results reach the same level as the best-reported for the given architecture when considering the multiple-discriminator setting.}
\label{tab:cifar32_scores}
\end{table*}

\begin{figure}[h]
\centering
\includegraphics[width=\columnwidth]{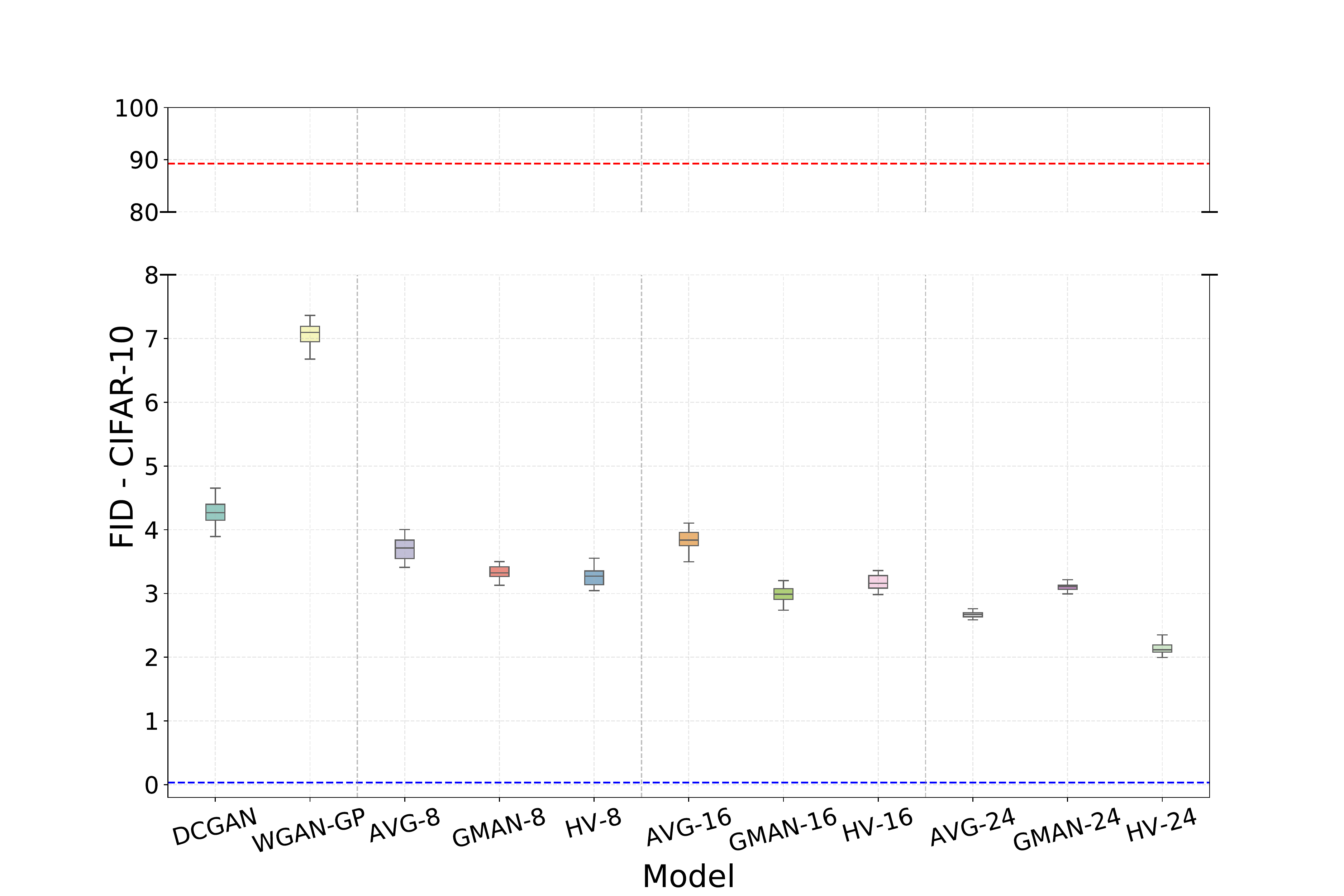}
\caption{Box-plots of 15 independent FID computations with $10000$ images. Dashed lines represent the FID for real data (blue) and a random generator (red). FID was computed with a pretrained ResNet.}
\label{fig:boxplot_cifar}
\end{figure}

\begin{figure}[h!]
\centering
\includegraphics[width=\columnwidth]{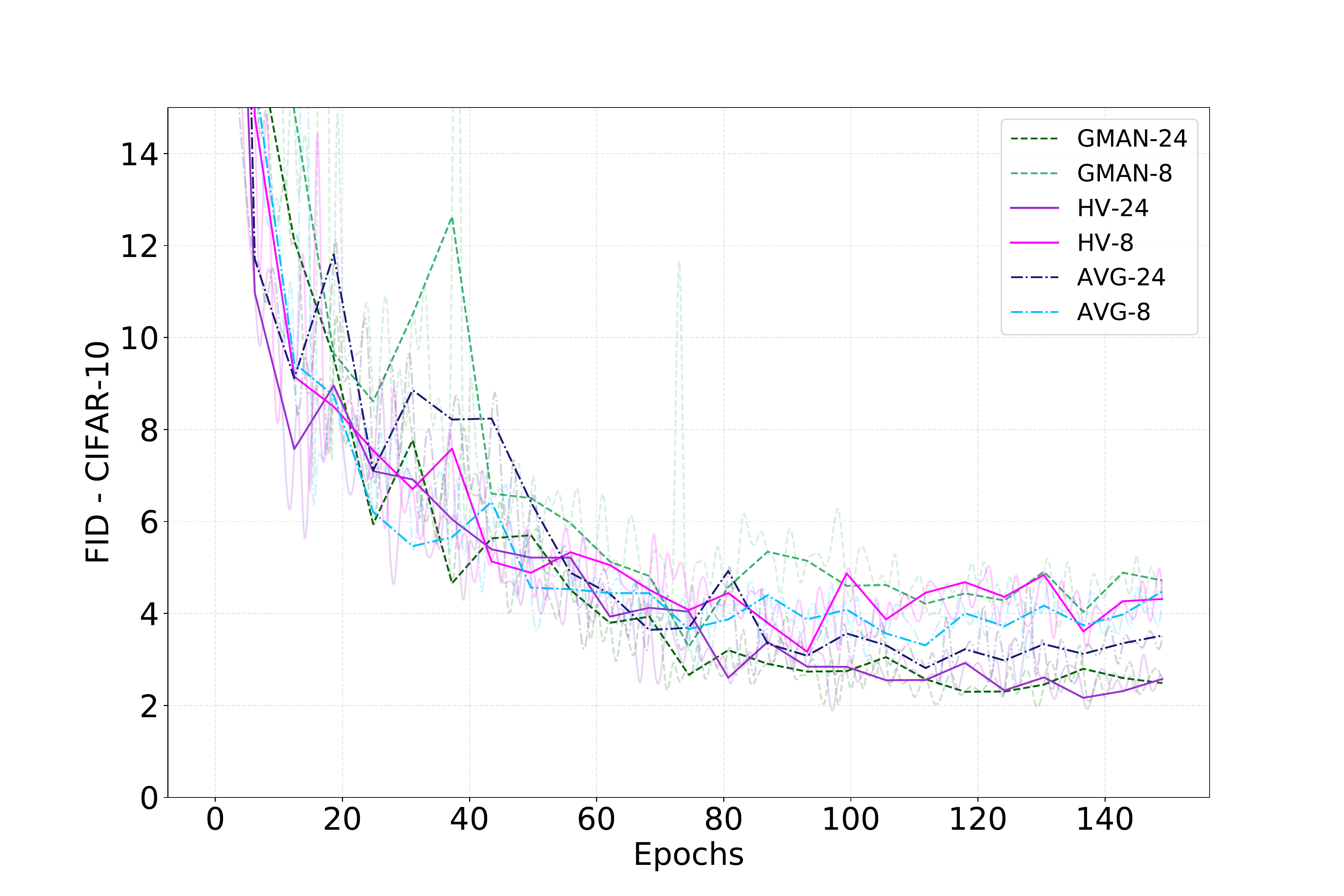}
\caption{FID estimated over $1000$ generated images at train time. Models trained against more discriminators achieve lower FID. FID was computed with a pretrained ResNet.}
\label{fig:fid_best_cifar}
\end{figure}

In Fig. \ref{fig:boxplot_cifar}, we report the box-plots of $15$ independent evaluations of FID on $10000$ images for the best model obtained with each method across $3$ independent runs. Results once more indicate that HV outperforms other methods in terms of quality of the generated samples. Moreover, performance clearly improves as the number of discriminators grows. Fig. \ref{fig:fid_best_cifar} shows the FID at train time, i.e. measured with $1000$ generated images after each epoch, for the best models across runs. Models trained against more discriminators clearly converge to smaller values. We report the norm of the update direction $||\sum_{k=1}^K \alpha_k \nabla l_k||$ for each method in Fig. \ref{fig:cifar_steep_time}-(a) in the Appendix.

\subsubsection{CIFAR-10}

We run experiments with CIFAR-10 in its original resolution aiming to contextualize our proposed approach with respect to previously introduced methods. We thus repeated similar experiments as reported in \cite{miyato2018spectral}-Table 2, for the model referred to as \emph{standard CNN}. The same architecture is employed and spectral normalization is removed from the discriminators, while a random projection input layer is added.

Results in terms of both FID and Inception score using their original implementations, evaluated on top of 5000 generated images as in \citep{miyato2018spectral} as well as with 10000 images, are reported in Table \ref{tab:cifar32_scores} for our proposed approach and our implementation of \citep{miyato2018spectral}, along with the FID measured using a ResNet classifier trained in advance on the CIFAR-10 dataset. 

As can be seen, the addition of the multiple discriminators setting along with hypervolume maximization yields a relevant shift in performance for the DCGAN-like generator, taking all evaluated metrics to within a competitive margin of recently proposed GANs, as well as outperforms our own implementation of SNGAN (using the best performing setup for this architecture as reported by \citet{miyato2018spectral}).

\subsection{Computational cost}


In Table \ref{tab:flops_mem} we present a comparison of minimum FID (measured with a pretrained ResNet) obtained during training, along with computation cost in terms of time and space for different GANs, with both 1 and 24 discriminators. The computational cost of training GANs under a multiple-discriminator setting is higher by design, in terms of both FLOPS and memory, if compared with single-discriminators settings. However, a corresponding improvement in performance is the result of the additional cost. This effect was consistently observed using 3 different well-known approaches, namely DCGAN \citep{radford2015unsupervised}, Least-square GAN (LSGAN) \citep{mao2017least}, and HingeGAN \citep{miyato2018spectral}. The architectures of all single discriminator models follow that of DCGAN, described in \citep{radford2015unsupervised}. For the 24 discriminators models, we used the setting described in Section \ref{sec:up_cifar}. All models were trained with minibatch of size 64 over 150 epochs.

We further highlight that even though training with multiple discriminators may be more computationally expensive when compared to conventional approaches, such a framework supports fully parallel training of the discriminators, a feature which is not trivially possible in other GAN settings. For example in WGAN, the discriminator is serially updated multiple times for each generator update. In Fig. \ref{fig:cifar_steep_time}-(b) in the Appendix, we provide a comparison between wall-clock time per iteration between all methods evaluated. Serial implementations of discriminator updates with 8 and 16 discriminators were observed to run faster than WGAN-GP. Moreover, all experiments performed within this work were executed in single GPU hardware, which indicates the multiple discriminator setting is a practical approach.

\begin{table}[h]
\centering
\resizebox{\columnwidth}{!}{
\begin{tabular}{ccccc}
\hline
                          & \# Disc. & FID-ResNet  & FLOPS & Memory  \\ \hline
\multirow{2}{*}{DCGAN}    & 1        & 4.22        & 8e10        & 1292            \\
                          & 24       & 1.89        & 5e11        & 5671            \\ \hline
\multirow{2}{*}{LSGAN}    & 1        & 4.55        & 8e10        & 1303            \\
                          & 24       & 1.91        & 5e11        & 5682            \\ \hline
\multirow{2}{*}{HingeGAN} & 1        & 6.17        & 8e10        & 1303            \\
                          & 24       & 2.25        & 5e11        & 5682            \\ \hline
\end{tabular}}
\caption{Comparison between different GANs with 1 and 24 discriminators in terms of minimum FID-ResNet obtained during training, and FLOPs (MAC) and memory consumption (MB) for a complete training step.}
\label{tab:flops_mem}
\end{table}

\subsection{Effect of the number of discriminators on sample diversity}
We repeat the experiments in \citep{srivastava2017veegan} aiming to analyze how the number of discriminators affects the sample diversity of the corresponding generator when trained using hypervolume maximization. The stacked MNIST dataset is employed and results reported in \citep{lin2017pacgan} are used for comparison. HV results for 8, 16, and 24 discriminators were obtained with 10k and 26k generator images, averaged over 10 runs. The number of covered modes along with the KL divergence between the generated mode distribution and test data are reported in Table \ref{tab:stck_mnist}.

\begin{table}
\vspace{0.65cm}
\centering
\resizebox{\columnwidth}{!}{
\begin{tabular}{ccc}
\hline
Model                                  & Modes (Max 1000)  & KL                   \\ \hline
DCGAN \citep{radford2015unsupervised}        & $99.0$            & $3.400$             \\
ALI \citep{dumoulin2016adversarially}        & $16.0$            & $5.400$             \\
Unrolled GAN \citep{metz2016unrolled}        & $48.7$            & $4.320$             \\
VEEGAN \citep{srivastava2017veegan}          & $150.0$           & $2.950$             \\
PacDCGAN2 \citep{lin2017pacgan}              & $1000.0 \pm 0.0$  & $0.060 \pm 0.003$   \\ \hline
HV - 8 disc. (10k)                           & $679.2 \pm 5.9$    & $1.139 \pm 0.011$   \\
HV - 16 disc. (10k)                          & $998.0 \pm 1.8$    & $0.120 \pm 0.004$   \\
HV - 24 disc. (10k)                          & $998.3 \pm 1.1$    & $0.116 \pm 0.003$   \\ \hline
HV - 8 disc. (26k)                           & $776.8 \pm 6.4$    & $1.115 \pm 0.007$   \\
HV - 16 disc. (26k)                          & $1000.0 \pm 0.0$   & $0.088 \pm 0.002$   \\
HV - 24 disc. (26k)                          & $1000.0 \pm 0.0$   & $0.084 \pm 0.002$   \\ \hline
\end{tabular}}
\caption{Number of covered modes and reverse KL divergence for stacked MNIST. We evaluate HV under a reduced test sample size (10k) with the goal of highlighting the effect provided by the increased number of discriminators on sample diversity.} 
\label{tab:stck_mnist}
\end{table}
As in previous experiments, results consistently improved as we increased the number of discriminators. All evaluated models using HV outperformed DCGAN, ALI, Unrolled GAN and VEEGAN. Moreover, HV with 16 and 24 discriminators achieved state-of-the-art coverage values. Thus, increasing each model's capacity by using more discriminators directly resulted in an improvement in the corresponding generator coverage. Training details as well as architecture information are presented in the Appendix.

\section{Conclusion}
\label{sec: conclusion}

In this work we show that employing multiple discriminators on GAN training is a practical approach for directly trading extra capacity - and thereby extra computational cost - for higher quality and diversity of generated samples. Such an approach is complimentary to other advances in GANs training and can be easily used together with other methods. We introduce a multi-objective optimization framework for studying multiple discriminator GANs, and showed strong similarities between previous work and the multiple gradient descent algorithm. The proposed approach was observed to consistently yield higher quality samples in terms of FID, and increasing the number of discriminators was shown to increase sample diversity and generator robustness.

Deeper analysis of the quantity $||\sum_{k=1}^K \alpha_k \nabla l_k||$ is a subject of future investigation. We hypothesize that using it as a penalty term might reduce the necessity of a high number of discriminators.

\bibliographystyle{icml2019.bst}

\bibliography{bibliography.bib}

\newpage

\onecolumn
\section*{\LARGE{Appendix}}

\section*{A - Objective evaluation metric.}

In \citep{heusel2017gans}, authors proposed to use as a quality metric the squared Fr{\'e}chet distance \citep{frechet1957distance} between Gaussians defined by estimates of the first and second order moments of the outputs obtained through a forward pass in a pretrained classifier of both real and generated data. They proposed the use of Inception V3 \citep{szegedy2016rethinking} for computation of the data representation and called the metric Fr{\'e}chet Inception Distance (FID), which is defined as:

\begin{equation}
\label{eq:fid_def}
\text{FID} = || m_d - m_g ||^2 + \text{Tr}(\Sigma_d+\Sigma_g-2(\Sigma_d \Sigma_g)^{\frac{1}{2}}),
\end{equation}
where $m_d, \Sigma_d$ and $m_g, \Sigma_g$ are estimates of the first and second order moments from the representations of real data distributions and generated data, respectively.

We employ FID throughout our experiments for comparison of different approaches. However, in datasets other than CIFAR-10 at its original resollution, for each dataset in which FID was computed, the output layer of a pretrained classifier on that particular dataset was used instead of Inception. $m_d$ and $\Sigma_d$ were estimated on the complete test partitions, which are not used during training.

\section*{B - Experimental setup for stacked MNIST experiments and generator's samples}

Architectures of the generator and discriminator are detailed in Tables \ref{tab:gen_arch} and \ref{tab:disc_arch}, respectively. Batch normalization was used in all intermediate convolutional and fully connected layers of both models. We employed RMSprop to train all the models with learning rate and $\alpha$ set to $0.0001$ and $0.9$, respectively. Mini-batch size was set to $64$. The setup in \citep{lin2017pacgan} is employed and we build 128000 and 26000 samples for train and test sets, respectively.  

\begin{table}[h]
\centering
\begin{tabular}{ccccc}
\hline
Layer                             & Outputs  & Kernel size & Stride & Activation \\ \hline
Input: $z \sim \mathcal{N}(0, I_{100})$ &          &             &        &            \\
Fully connected                   & 2*2*512  & 4, 4        & 2, 2   & ReLU       \\
Transposed convolution            & 4*4*256  & 4, 4        & 2, 2   & ReLU       \\
Transposed convolution            & 8*8*128  & 4, 4        & 2, 2   & ReLU       \\
Transposed convolution            & 14*14*64 & 4, 4        & 2, 2   & ReLU       \\
Transposed convolution            & 28*28*3  & 4, 4        & 2, 2   & Tanh       \\ \hline
\end{tabular}
\vspace{0.1cm}
\caption{Generator's architecture.}
\label{tab:gen_arch}
\end{table}

\begin{table}[h]
\centering
\begin{tabular}{ccccc}
\hline
Layer       & Outputs & Kernel size & Stride & Activation \\ \hline
Input       & 28*28*3 &             &        &            \\
Projection  & 14*14*3 & 8, 8        & 2, 2   &            \\
Convolution & 7*7*64  & 4,4         & 2, 2   & LeakyReLU  \\
Convolution & 5*5*128 & 4, 4        & 2, 2   & LeakyReLU  \\
Convolution & 2*2*256 & 4, 4        & 2, 2   & LeakyReLU  \\
Convolution & 1       & 4, 4        & 2, 2   & Sigmoid    \\ \hline
\end{tabular}
\vspace{0.1cm}
\caption{Discriminator's architecture.}
\label{tab:disc_arch}
\end{table}

\begin{figure}[H]
    \centering
    \begin{subfigure}[]{\textwidth}
        \centering
        \includegraphics[width=0.68\textwidth]{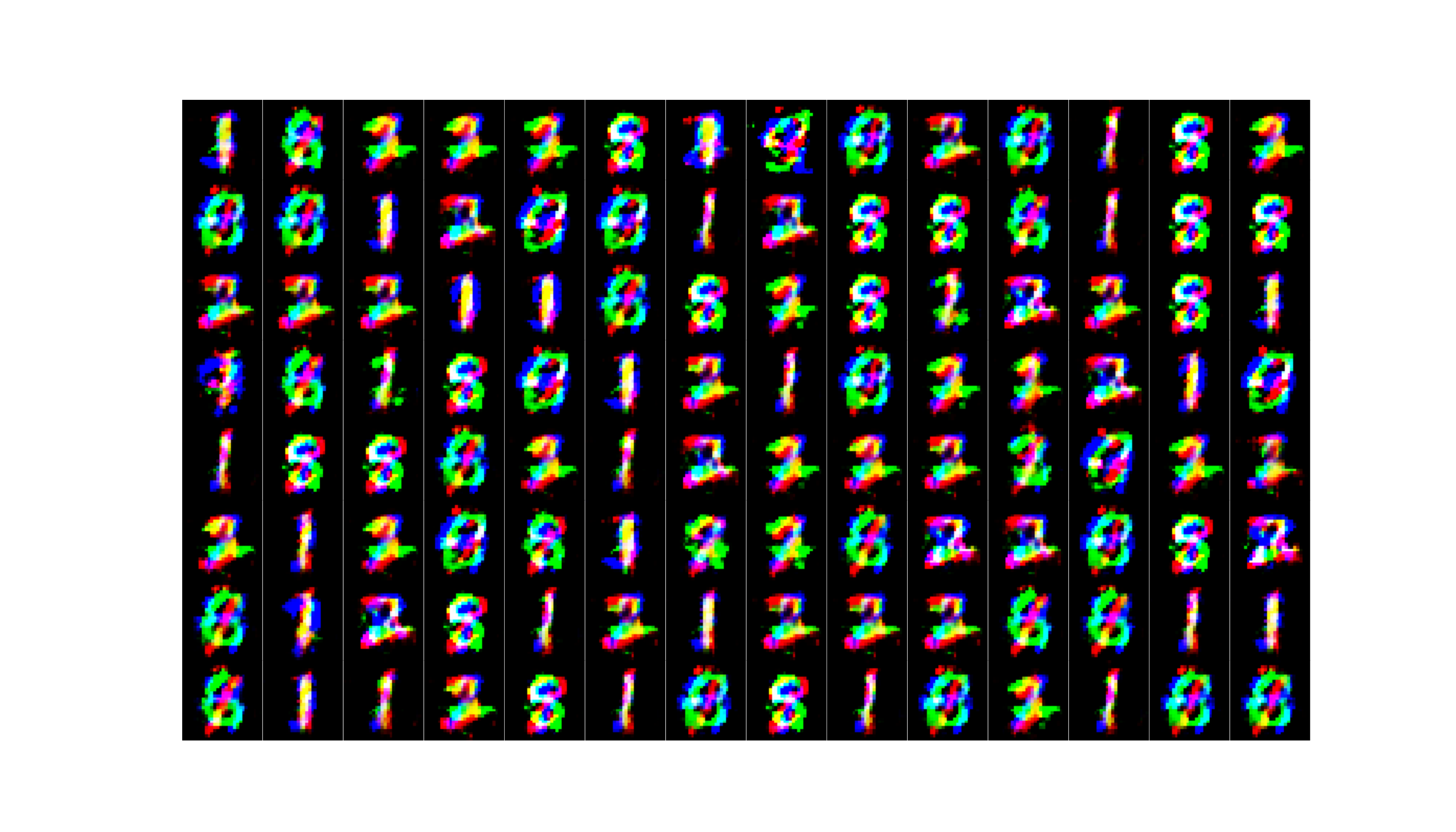}
        \caption{HV - 8 discriminators}
    \end{subfigure}%
    \\
    \begin{subfigure}[]{\textwidth}
        \centering
        \includegraphics[width=0.68\textwidth]{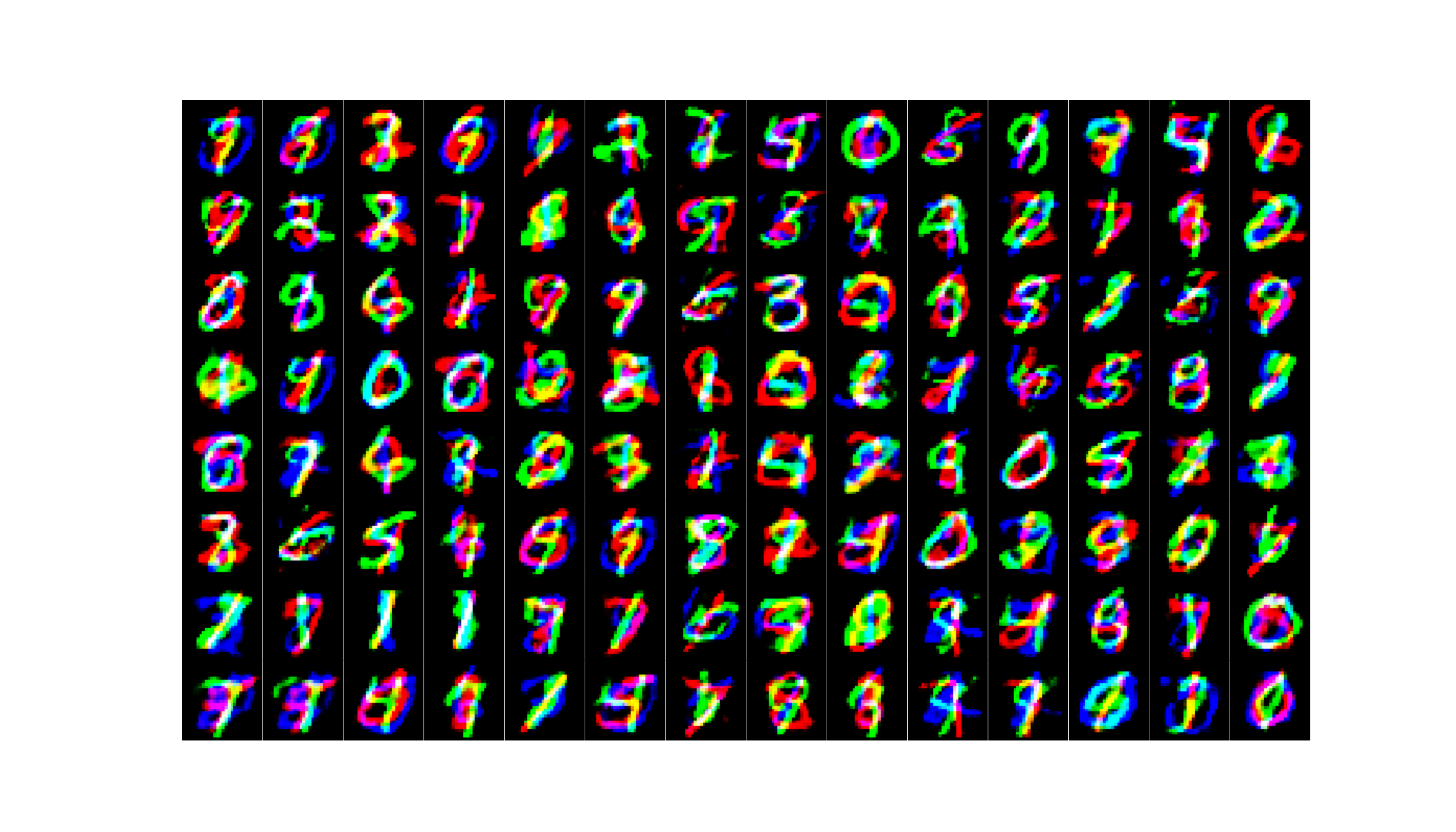}
        \caption{HV - 16 discriminators}
    \end{subfigure}
    \\
    \begin{subfigure}[]{\textwidth}
        \centering
        \includegraphics[width=0.68\textwidth]{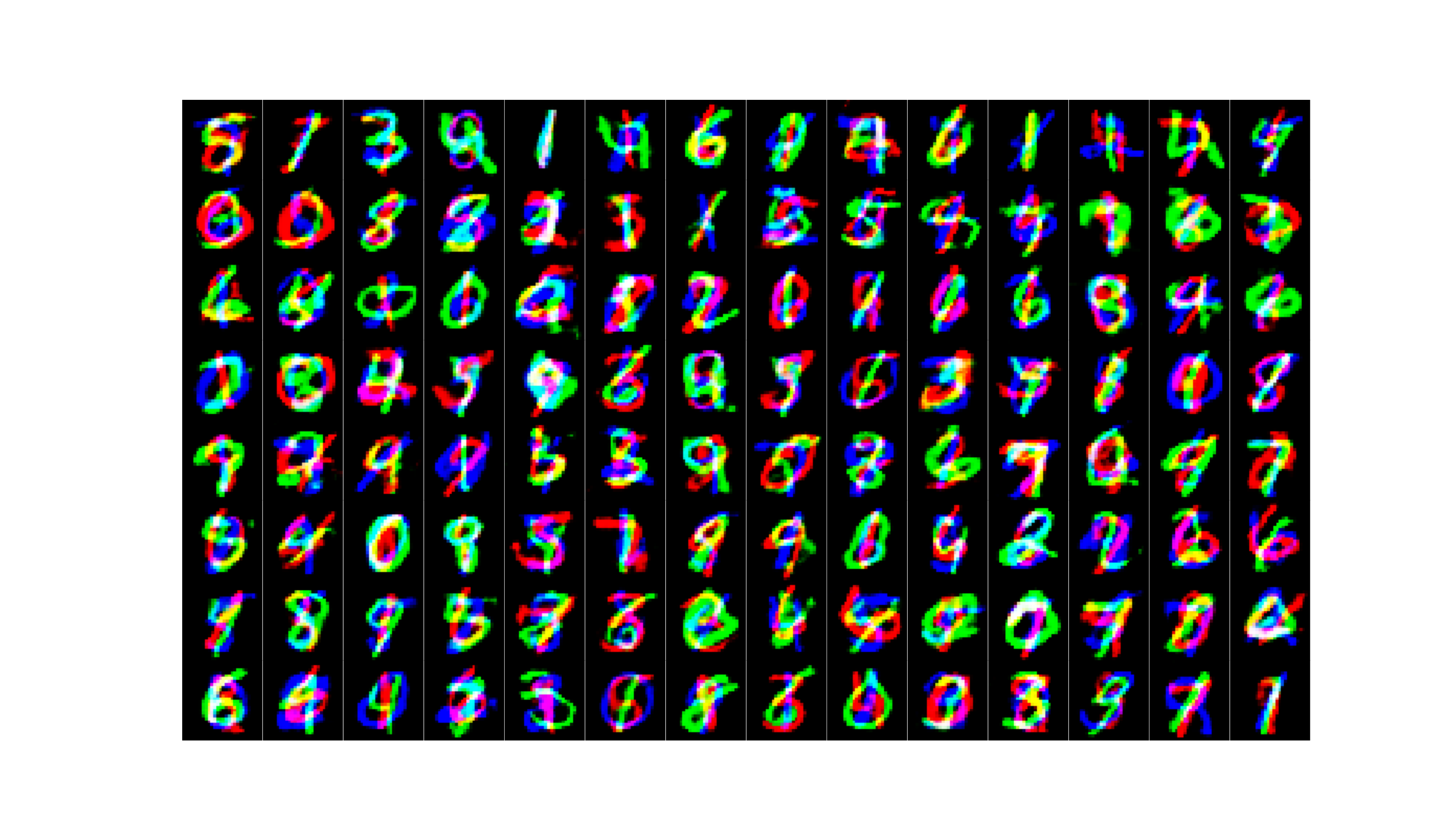}
        \caption{HV - 24 discriminators}
    \end{subfigure}%
     \caption{Stacked MNIST samples for HV trained with 8, 16, and 24 discriminators. Samples diversity increases greatly when more discriminators are employed.}
		\label{fig:stkd_mnist}
\end{figure}

\newpage
\section*{C - Extra results on upscaled CIFAR-10}

\subsection*{C.1 - Multiple discriminators across different initializations and other scores}

Table \ref{tab:fid-cifar} presents the best FID (computed with a pretrained ResNet) achieved by each approach at train time, along with the epoch in which it was achieved, for each of 3 independent runs. Train time FIDs are computed using 1000 generated images.

\begin{table}[h]
\centering
\resizebox{0.4\textwidth}{!}{
\begin{tabular}{ccc}
\hline
\textbf{\#D}        & \textbf{Method} & \textbf{Best FID (epoch)}          \\ \hline
\multirow{2}{*}{1}  & DCGAN           & 7.09 (68), 9.09 (21), 4.22 (101)   \\
                    & WGAN-GP         & 5.09 (117), 5.69 (101) 7.13 (71)    \\ \hline
\multirow{3}{*}{8}  & AVG             & 3.35 (105), 4.64 (141), 3.00 (76)  \\
                    & GMAN            & 4.28 (123), 4.24 (129), 3.80 (133) \\
                    & HV              & 3.87 (102), 4.54 (82), 3.20 (98)   \\ \hline
\multirow{3}{*}{16} & AVG             & 3.16 (96), 2.50 (91), 2.77 (116)   \\
                    & GMAN            & 2.69 (129), 2.36 (144), 2.48 (120) \\
                    & HV              & 2.56 (85), 2.70 (97), 2.68 (133)   \\ \hline
\multirow{3}{*}{24} & AVG             & 2.10 (94), 2.44 (132), 2.43 (129) \\
                    & GMAN            & 2.16 (120), 2.02 (98), 2.13 (130)  \\
                    & HV              & 2.05 (83), 1.89 (97), 2.23 (130)  \\ \hline
\end{tabular}}
\caption{Best FID obtained for each approach on $3$ independent runs. FID is computed on $1000$ generated images after every epoch.}
\label{tab:fid-cifar}
\end{table}

\begin{figure}[H]
    \centering
    \begin{subfigure}[htbp]{0.4\textwidth}
        \centering
        \includegraphics[width=1.\textwidth]{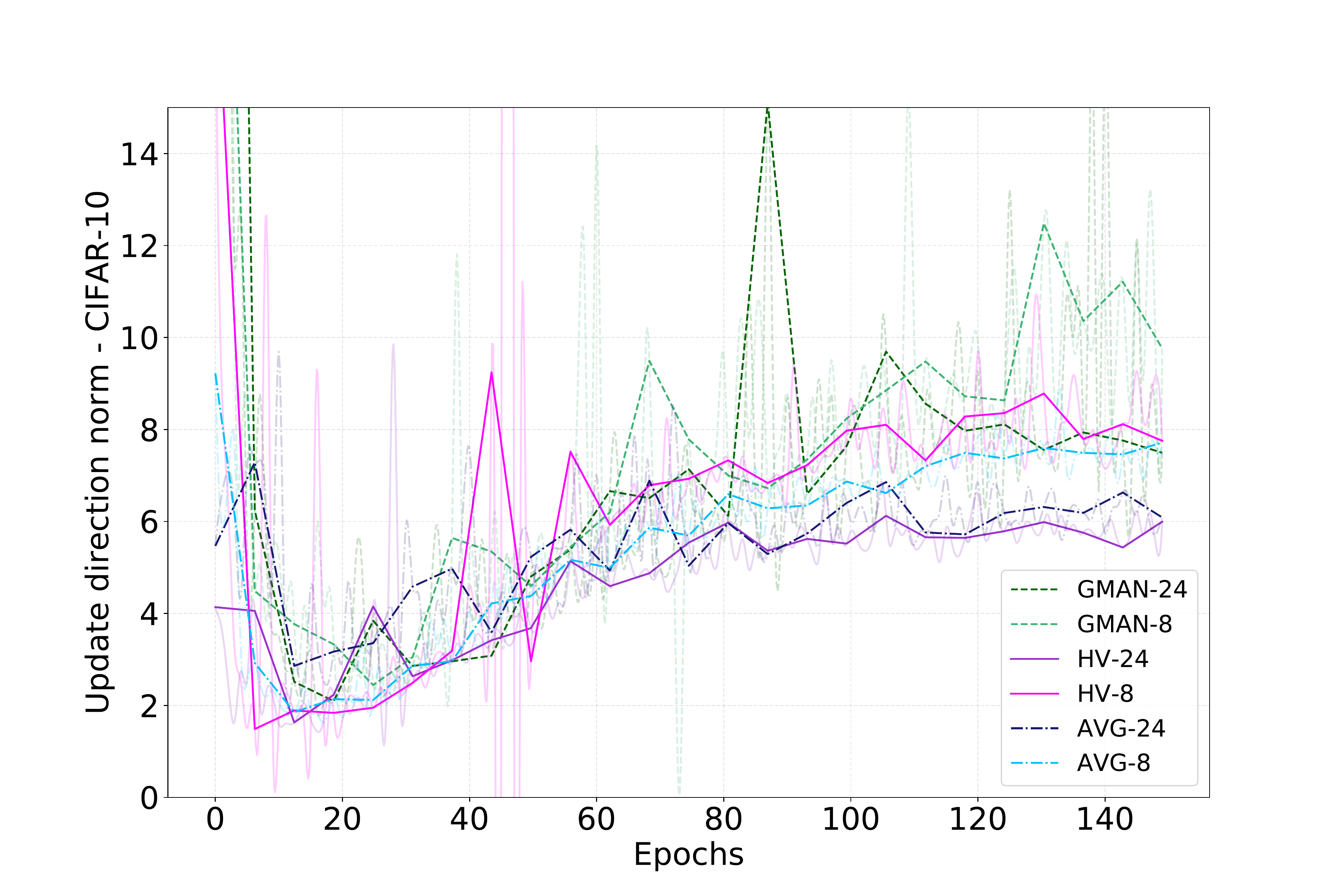}
        \caption{Norm of the update direction over time for each method. Higher number of discriminators yield lower norm upon convergence.}
    \end{subfigure}%
    \hspace{0.7cm}
    \begin{subfigure}[htbp]{0.4\textwidth}
        \centering
        \includegraphics[width=1.\textwidth]{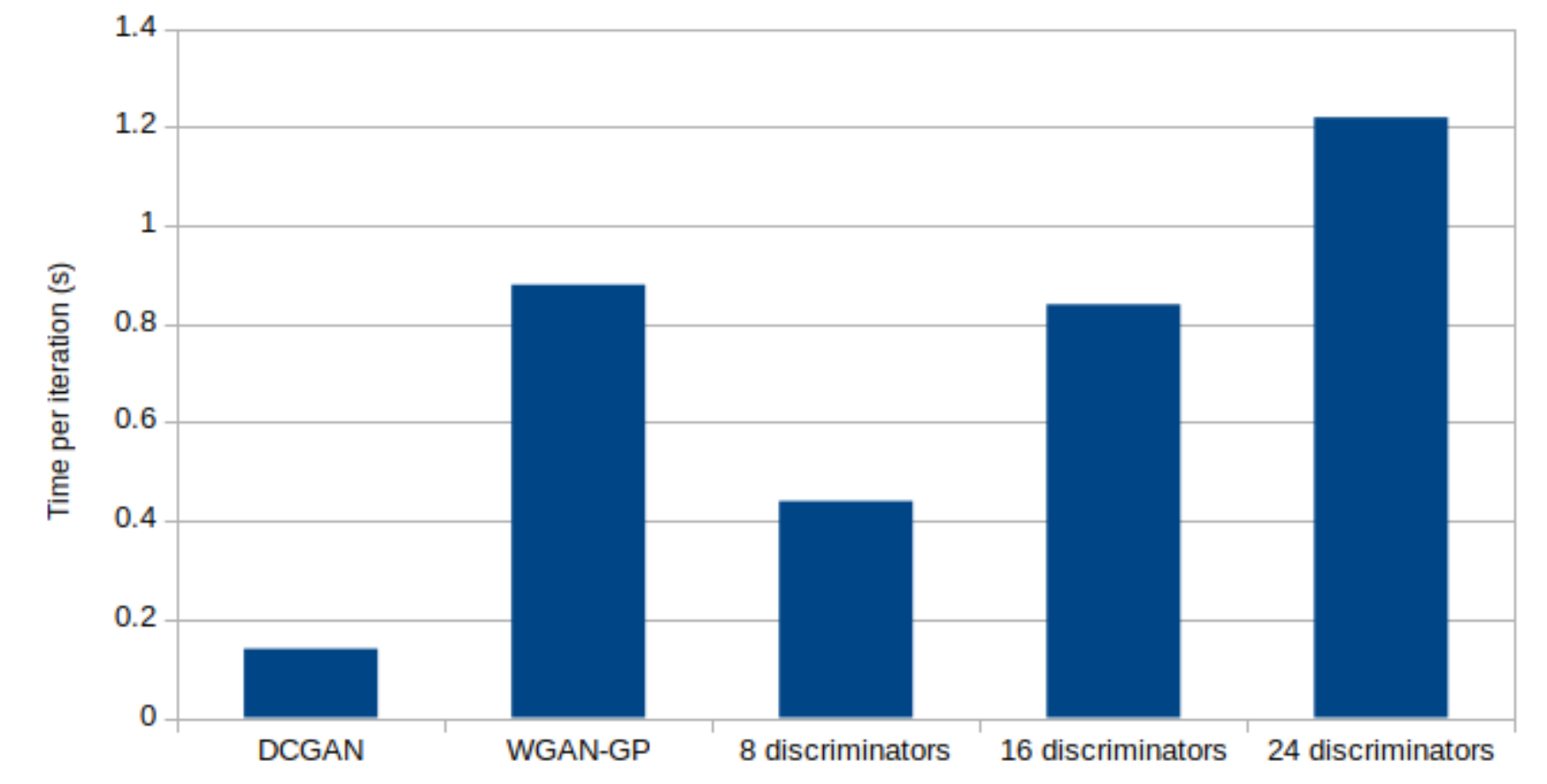}
        \caption{Time in seconds per iteration of each method for serial updates of discriminators. The different multiple discriminators approaches considered do not present relevant difference in time per iteration.}
    \end{subfigure}
    \caption{(a) Norm of update direction. (b) Time per iteration.}
    \label{fig:cifar_steep_time}
\end{figure}

In Fig. \ref{fig:cifar_steep_time}-(a), we report the norm of the update direction $||\sum_{k=1}^K \alpha_k \nabla l_k||$ of the best model obtained for each method. Interestingly, different methods present similar behavior in terms of convergence in the Pareto-stationarity sense, i.e. the norm upon convergence is lower for models trained against more discriminators, regardless of the employed method.



We computed extra scores using 10000 images generated by the best model reported in Table \ref{tab:fid-cifar}, i.e. the same models utilized to generate the results shown in Fig. \ref{fig:boxplot_cifar}. Both Inception score and FID were computed with original implementations, while FID-VGG and FID-ResNet were computed using a VGG and a ResNet we pretrained. Results are reported with respect to DCGAN's scores to avoid direct comparison with results reported elsewhere for CIFAR-10 on its usual resolution ($32\times32$).

\begin{table}[h]
\centering
\resizebox{0.8\textwidth}{!}{
\begin{tabular}{ccccccccccc}
\hline
\textit{\textbf{}}       & \textbf{WGAN-GP} & \textbf{AVG-8} & \textbf{AVG-16} & \textbf{AVG-24} & \textbf{GMAN-8} & \textbf{GMAN-16} & \textbf{GMAN-24} & \textbf{HV-8} & \textbf{HV-16} & \textbf{HV-24} \\ \hline
\textit{Inception Score} & 1.08             & 1.02           & 1.26            & 1.36            & 0.95            & 1.32             & 1.42             & 1.00          & 1.30           & \textbf{1.44}  \\ \hline
\textit{FID}             & 0.80             & 0.98           & 0.76            & 0.73            & 0.92            & 0.79             & \textbf{0.65}    & 0.89          & 0.77           & 0.72           \\
\textit{FID-VGG}         & 1.29             & 0.91           & 1.03            & 0.85            & 0.87            & 0.78             & 0.73             & 0.78          & 0.75           & \textbf{0.64}  \\
\textit{FID-ResNet}      & 1.64             & 0.88           & 0.90            & 0.62            & 0.80            & 0.72             & 0.73             & 0.75          & 0.73           & \textbf{0.51}  \\ \hline
\end{tabular}}
\caption{Scores of different methods measure on generated samples from the upsacaled CIFAR-10. DCGAN scores are used as reference values, and results report are the ratio between given model and DCGAN scores. Inception score is better when high, whereas FIDs are better when low.}
\label{tab:scores_cifar}
\end{table}

\newpage
\section*{D - CelebA dataset 128x128}
\begin{figure}[H]
    \centering
    \begin{subfigure}[htbp]{\textwidth}
    \centering
    \begin{tabular}{cccc}
    \includegraphics[width = 1in]{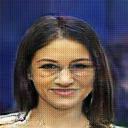}&
    \includegraphics[width = 1in]{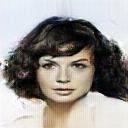} &
    \includegraphics[width = 1in]{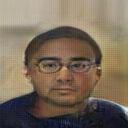} &
    \includegraphics[width = 1in]{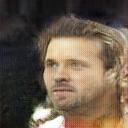} \\
    \includegraphics[width = 1in]{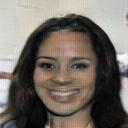} &
    \includegraphics[width = 1in]{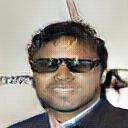} &
    \includegraphics[width = 1in]{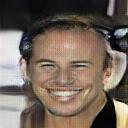} &
    \includegraphics[width = 1in]{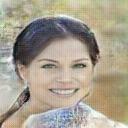} \\
    \end{tabular}
    \caption{HV - 6 discriminators}
    \end{subfigure}%
    \\
    \begin{subfigure}[htbp]{\textwidth}
    \centering
    \begin{tabular}{cccc}
    \includegraphics[width = 1in]{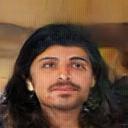}&
    \includegraphics[width = 1in]{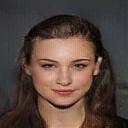} &
    \includegraphics[width = 1in]{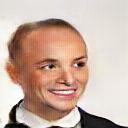} &
    \includegraphics[width = 1in]{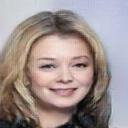} \\
    \includegraphics[width = 1in]{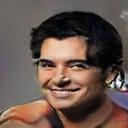} &
    \includegraphics[width = 1in]{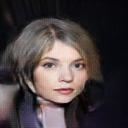} &
    \includegraphics[width = 1in]{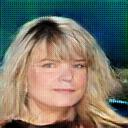} &
    \includegraphics[width = 1in]{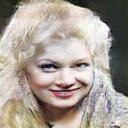} \\
    \end{tabular}
    \caption{HV - 8 discriminators}
    \end{subfigure}
    \\
    \begin{subfigure}[htbp]{\textwidth}
    \centering
    \begin{tabular}{cccc}
    \includegraphics[width = 1in]{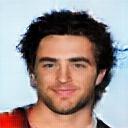}&
    \includegraphics[width = 1in]{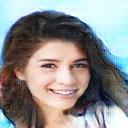} &
    \includegraphics[width = 1in]{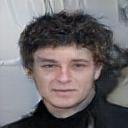} &
    \includegraphics[width = 1in]{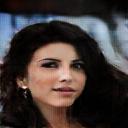} \\
    \includegraphics[width = 1in]{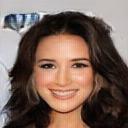} &
    \includegraphics[width = 1in]{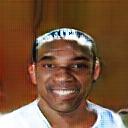} &
    \includegraphics[width = 1in]{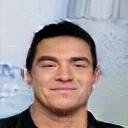} &
    \includegraphics[width = 1in]{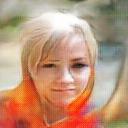} \\
    \end{tabular}
    \caption{HV - 10 discriminators}
    \end{subfigure}%
  \caption{128x128 CelebA samples for HV trained during 24 epochs with 6, 8, and 10 discriminators.} 
  \label{fig:celeba_128}
\end{figure}

In this experiment, we verify whether the proposed multiple discriminators setting is capable of generating higher resolution images. For that, we employed the CelebA at a size of 128x128. We used a similar architecture for both generator and discriminators networks as described in the previous experiments. A convolutional layer with 2048 feature maps was added to both generator and discriminators architectures due to the increase in the image size. Adam optimizer with the same set of hyperparameters as for CIFAR-10 and CelebA 64x64 was employed. We trained models with 6, 8, and 10 discriminators during 24 epochs. Samples from each generator are shown in Figure \ref{fig:celeba_128}.  

\clearpage

\section*{E - Generating 256x256 Cats}
\begin{figure}[h]
\centering
\begin{tabular}{ccc}
\includegraphics[width = 1.2in]{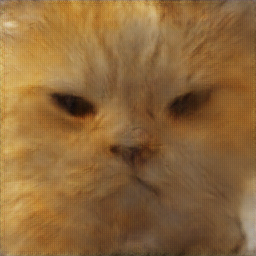}&
\includegraphics[width = 1.2in]{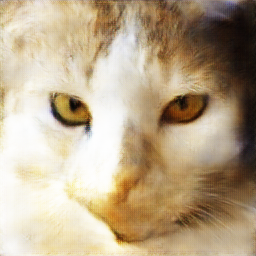} &
\includegraphics[width = 1.2in]{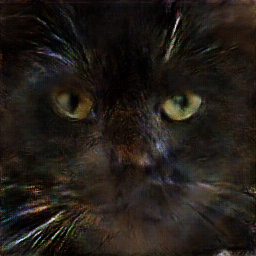} \\
\includegraphics[width = 1.2in]{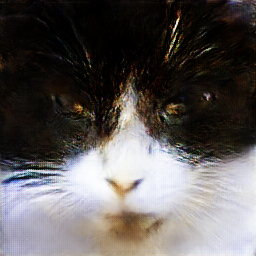} &
\includegraphics[width = 1.2in]{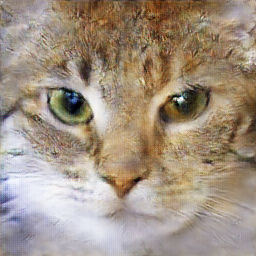} &
\includegraphics[width = 1.2in]{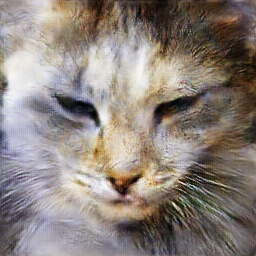}\\
\includegraphics[width = 1.2in]{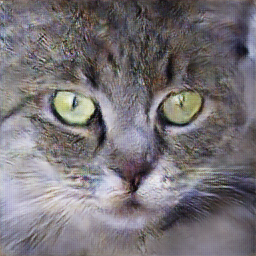} &
\includegraphics[width = 1.2in]{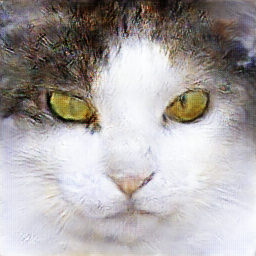} &
\includegraphics[width = 1.2in]{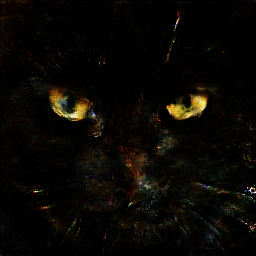} \\
\includegraphics[width = 1.2in]{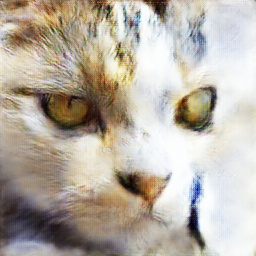} &
\includegraphics[width = 1.2in]{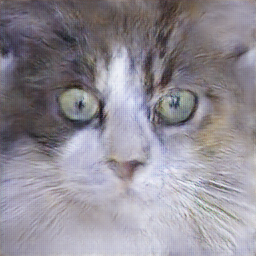} &
\includegraphics[width = 1.2in]{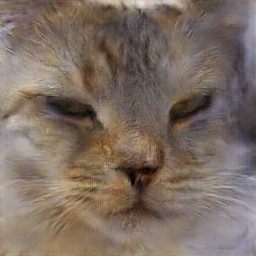} \\
\includegraphics[width = 1.2in]{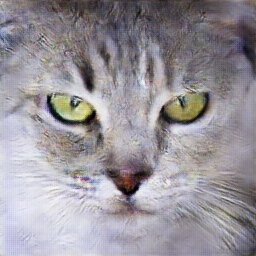} &
\includegraphics[width = 1.2in]{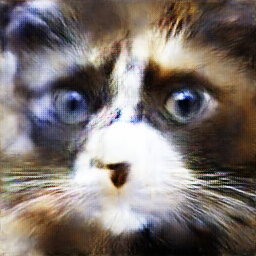} &
\includegraphics[width = 1.2in]{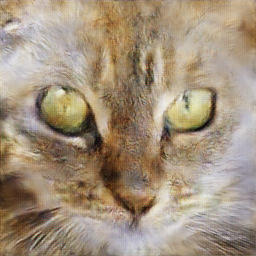} \\
\end{tabular}
\caption{Cats generated using 24 discriminators after 288 training epochs.}
\label{fig:cats}
\end{figure}

We show the proposed multiple-discriminators setting scales to higher resolution even in the small dataset regime, by reproducing the experiments presented in \citep{jolicoeur2018relativistic}. We used the same architecture for the generator. For the discriminator, we removed batch normalization from all layers and used stride equal to 1 at the last convolutional layer, after adding the initial projection step. The Cats dataset \footnote{https://www.kaggle.com/crawford/cat-dataset} was employed, we followed the same pre-processing steps, which, in our case, yielded 1740  training samples with resolution of 256x256. Our model is trained using 24 discriminators and Adam optimizer with the same hyperparameters as for CIFAR-10 and CelebA previously described experiments. In Figure \ref{fig:cats} we show generator's samples after 288 training epochs. One epoch corresponds to updating over 27 minibatches of size 64.

\clearpage

\section*{H - Wall-clock time for reaching best FID during training on MNIST}

\end{document}